\pdfoutput=1

\documentclass[11pt]{article}

\usepackage[preprint]{acl}

\usepackage{times}
\usepackage{latexsym}

\usepackage[T1]{fontenc}

\usepackage[utf8]{inputenc}

\usepackage{microtype}

\usepackage{inconsolata}

\usepackage{graphicx}

\usepackage{amsmath}
\usepackage{amsthm}
\usepackage{hyperref}       
\usepackage{url}            
\usepackage{booktabs}       
\usepackage{amsfonts}       
\usepackage{nicefrac}       
\usepackage{microtype}      
\usepackage{xcolor}         
\usepackage{subcaption}
\usepackage{rotating}
\usepackage[shortlabels]{enumitem}\setlist[enumerate, 1]{1.}
\usepackage{listings}

\newtheorem{lemma}{Lemma}

\newcommand\topk{\operatorname{top-k}}

\newcommand\attnscore{\operatorname{AttentionScore}}
\newcommand\attneig{\operatorname{AttnEigvals}}
\newcommand\lapeig{\operatorname{LapEigvals}}
\newcommand\attnlogdet{\operatorname{AttnLogDet}}
\newcommand\hiddenstate{\operatorname{HiddenStates}}

\newcommand\diag[1]{\operatorname{diag\left(#1\right)}}
\newcommand\sort[1]{\operatorname{sort\left(#1\right)}}
\DeclareMathOperator*{\concat}{%
    \mathchoice%
        {\Big\Vert}%
        {\big\Vert}%
        {\Vert}%
        {\Vert}%
}

\definecolor{aliceblue}{rgb}{0.94, 0.97, 1.0}

\lstdefinestyle{prompt}{
    basicstyle=\small\ttfamily,
    backgroundcolor=\color{aliceblue},
    frame=single,
    framesep=5pt,
    breaklines=true,
    postbreak=\mbox{\textcolor{red}{$\hookrightarrow$}\space},
}

\DeclareGraphicsExtensions{.pdf}

\title{Hallucination Detection in LLMs Using Spectral Features of Attention Maps}

\author{
 \textbf{Jakub Binkowski\textsuperscript{1}},
 \textbf{Denis Janiak\textsuperscript{1}},
 \textbf{Albert Sawczyn\textsuperscript{1}}\\
 \textbf{Bogdan Gabrys\textsuperscript{2}},
 \textbf{Tomasz Kajdanowicz\textsuperscript{1}}
\\
 \textsuperscript{1}Wroclaw University of Science and Technology,
 \textsuperscript{2}University of Technology Sydney,
\\
\small{
   \textbf{Correspondence:} \href{mailto:jakub.binkowski@pwr.edu.pl}{jakub.binkowski@pwr.edu.pl}
 }
}

\begin{document}
\newcommand{\llamabig}{\texttt{Llama-3.1-8B}}
\newcommand{\llamasmall}{\texttt{Llama-3.2-3B}}
\newcommand{\mistralnemo}{\texttt{Mistral-Nemo}}
\newcommand{\mistralsmall}{\texttt{Mistral-Small-24B}}
\newcommand{\philllm}{\texttt{Phi-3.5}}
\newcommand{\llmjudge}{\textit{llm-as-judge}}
\newcommand{\gptmini}{\texttt{gpt-4o-mini}}
\newcommand{\gptnew}{\texttt{gpt-4.1}}

\maketitle

\begin{abstract}
Large Language Models (LLMs) have demonstrated remarkable performance across various tasks but remain prone to hallucinations. Detecting hallucinations is essential for safety-critical applications, and recent methods leverage attention map properties to this end, though their effectiveness remains limited. In this work, we investigate the spectral features of attention maps by interpreting them as adjacency matrices of graph structures. We propose the $\lapeig$ method, which utilizes the top-$k$ eigenvalues of the Laplacian matrix derived from the attention maps as an input to hallucination detection probes. Empirical evaluations demonstrate that our approach achieves state-of-the-art hallucination detection performance among attention-based methods. Extensive ablation studies further highlight the robustness and generalization of $\lapeig$, paving the way for future advancements in the hallucination detection domain.
\end{abstract}

\section{Introduction}
The recent surge of interest in Large Language Models (LLMs), driven by their impressive performance across various tasks, has led to significant advancements in their training, fine-tuning, and application to real-world problems. Despite progress, many challenges remain unresolved, particularly in safety-critical applications with a high cost of errors. A significant issue is that LLMs are prone to hallucinations, i.e. generating "content that is nonsensical or unfaithful to the provided source content" \citep{farquhar_detecting_2024, huang_survey_2023}. Since eliminating hallucinations is impossible \citep{lee_mathematical_2023,xu_hallucination_2024}, there is a pressing need for methods to detect when a model produces hallucinations. In addition, examining the internal behavior of LLMs in the context of hallucinations may yield important insights into their characteristics and support further advancements in the field. Recent studies have shown that hallucinations can be detected using internal states of the model, e.g., hidden states \citep{chen_inside_2024} or attention maps \citep{chuang_lookback_2024}, and that LLMs can internally "know when they do not know" \citep{azaria_internal_2023, orgad_llms_2025}. We show that spectral features of attention maps coincide with hallucinations and, building on this observation, propose a novel method for their detection.

As highlighted by \citep{barbero_transformers_2024}, attention maps can be viewed as weighted adjacency matrices of graphs. Building on this perspective, we performed statistical analysis and demonstrated that the eigenvalues of a Laplacian matrix derived from attention maps serve as good predictors of hallucinations. We propose the $\lapeig$ method, which utilizes the top-$k$ eigenvalues of the Laplacian as input features of a probing model to detect hallucinations. We share full implementation in a public repository: \url{https://github.com/graphml-lab-pwr/lapeigvals}.

We summarize our contributions as follows:
\begin{enumerate}[(1)]
    \itemsep0em 
    \item We perform statistical analysis of the Laplacian matrix derived from attention maps and show that it could serve as a better predictor of hallucinations compared to the previous method relying on the log-determinant of the maps.
    \item Building on that analysis and advancements in the graph-processing domain, we propose leveraging the top-$k$ eigenvalues of the Laplacian matrix as features for hallucination detection probes and empirically show that it achieves state-of-the-art performance among attention-based approaches.
    \item Through extensive ablation studies, we demonstrate properties, robustness and generalization of $\lapeig$ and suggest promising directions for further development.
\end{enumerate}

\section{Motivation}
\label{sec:motivation}

\begin{figure*}[!htb]
    \centering
    \includegraphics[width=\textwidth]{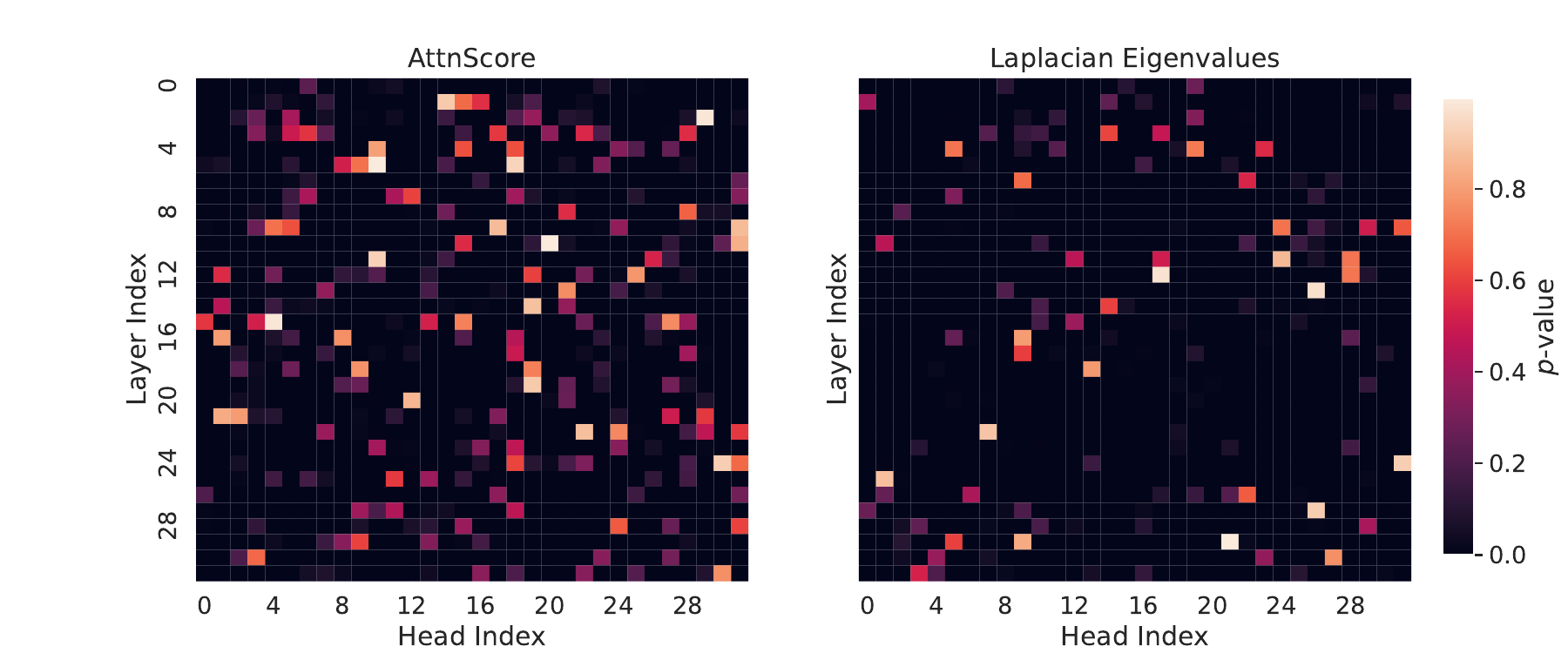}
    \caption{Visualization of $p$-values from the two-sided Mann-Whitney U test for all layers and heads of \llamabig\ across two feature types: $\attnscore$ and the $k{=}10$ Laplacian eigenvalues. These features were derived from attention maps collected when the LLM answered questions from the TriviaQA dataset. Higher $p$-values indicate no significant difference in feature values between hallucinated and non-hallucinated examples. For $\attnscore$, $80\%$ of heads have $p<0.05$, while for Laplacian eigenvalues, this percentage is $91\%$. Therefore, Laplacian eigenvalues may be better predictors of hallucinations, as feature values across more heads exhibit statistically significant differences between hallucinated and non-hallucinated examples.}
    \label{fig:motivation}
\end{figure*}

\begin{figure}[!htb]
    \centering
    \includegraphics[width=0.9\linewidth]{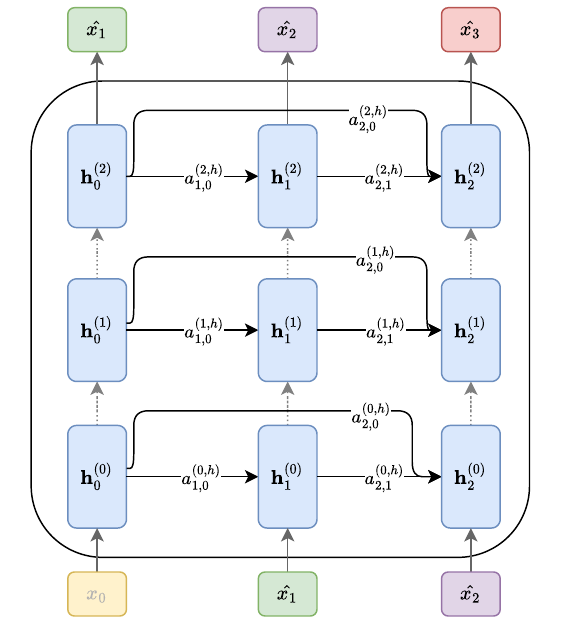}
    \caption{The autoregressive inference process in an LLM is depicted as a graph for a single attention head $h$ (as introduced by \citep{vaswani_attention_2017}) and three generated tokens ($\hat{x}_1, \hat{x}_2, \hat{x}_3$). Here, $\mathbf{h}^{(l)}_{i}$ represents the hidden state at layer $l$ for the input token $i$, while $a^{(l, h)}_{i, j}$ denotes the scalar attention score between tokens $i$ and $j$ at layer $l$ and attention head $h$. Arrows direction refers to information flow during inference.}
    \label{fig:graph}
\end{figure}

Considering the attention matrix as an adjacency matrix representing a set of Markov chains, each corresponding to one layer of an LLM \citep{NEURIPS2024_1ac3030f} (see Figure \ref{fig:graph}), we can leverage its spectral properties, as was done in many successful graph-based methods \citep{hahn_applications_1997,von_luxburg_tutorial_2007,bruna_spectral_2013, topping_understanding_2022}. In particular, it was shown that the graph Laplacian might help to describe several graph properties, like the presence of bottlenecks \citep{topping_understanding_2022, black_understanding_2023}. We hypothesize that hallucinations may arise from disruptions in information flow, such as bottlenecks, which could be detected through the graph Laplacian.

To assess whether our hypothesis holds, we computed graph spectral features and verified if they provide a stronger coincidence with hallucinations than the previous attention-based method - $\attnscore$ \citep{sriramanan_llm-check_2024}. We prompted an LLM with questions from the TriviaQA dataset \citep{joshi_triviaqa_2017} and extracted attention maps, differentiating by layers and heads. We then computed the spectral features, i.e., the 10 largest eigenvalues of the Laplacian matrix from each head and layer. Further, we conducted a two-sided Mann-Whitney U test \citep{mann1947test} to compare whether Laplacian eigenvalues and the values of $\attnscore$ are different between hallucinated and non-hallucinated examples. Figure~\ref{fig:motivation} shows $p$-values for all layers and heads, indicating that $\attnscore$ often results in higher $p$-values compared to Laplacian eigenvalues. Overall, we studied 7 datasets and 5 LLMs and found similar results (see Appendix~\ref{sec:appendix_stat_test}). Based on these findings, we propose leveraging top-$k$ Laplacian eigenvalues as features for a hallucination probe.

\section{Method}
\label{sec:method}

\begin{figure*}[!htb]
    \centering
    \includegraphics[width=\textwidth, keepaspectratio]{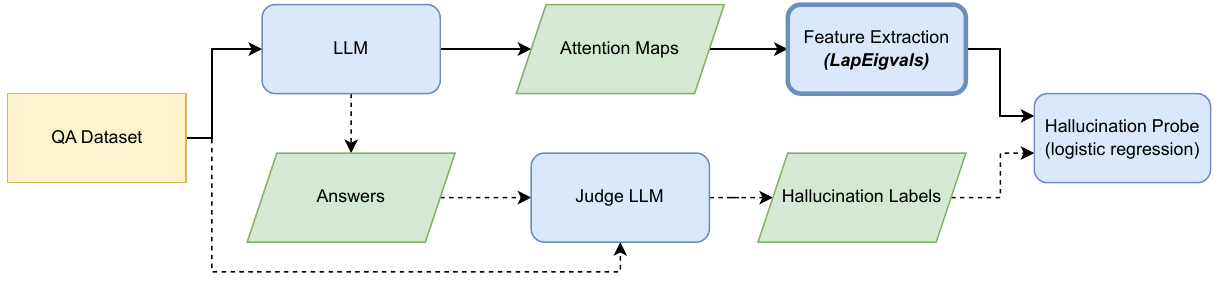}
    \caption{Overview of the methodology used in this work. Solid lines indicate the test-time pipeline, while dashed lines represent additional pipeline steps for generating labels for training the hallucination probe (logistic regression). The primary contribution of this work is leveraging the top-$k$ eigenvalues of the Laplacian as features for the hallucination probe, highlighted with a bold box on the diagram.}
    \label{fig:methodology}
\end{figure*}

In our method, we train a hallucination probe using only attention maps, which we extracted during LLM inference, as illustrated in Figure \ref{fig:graph}. The attention map is a matrix containing attention scores for all tokens processed during inference, while the hallucination probe is a logistic regression model that uses features derived from attention maps as input. This work's core contribution is using the top-$k$ eigenvalues of the Laplacian matrix as input features, which we detail below.

Denote $\mathbf{A}^{(l, h)} \in \mathbb{R}^{T \times T}$ as the attention map matrix for layer $l \in \{1\dotsc L\}$ and attention head $h \in \{1 \dotsc H\}$, where $T$ is the total number of tokens generated by an LLM (including input tokens), $L$ the number of layers (transformer blocks), and $H$ the number of attention heads. The attention matrix is row-stochastic, meaning each row sums to 1 ($\sum_{j=0}^{T} \mathbf{A}^{(l, h)}_{:,j} = \mathbf{1}$). It is also lower triangular ($a^{(l, h)}_{ij} = 0$ for all $j > i$) and non-negative ($a^{(l, h)}_{ij} \geq 0$ for all $i, j$). We can view $\mathbf{A}^{(l, h)}$ as a weighted adjacency matrix of a directed graph, where each node represents processed token, and each directed edge from token $i$ to token $j$ is weighted by the attention score, as depicted in Figure~\ref{fig:graph}.

Then, we define the Laplacian of a layer $l$ and attention head $h$ as:
\begin{equation}
\label{eq:laplacian}
    \mathbf{L}^{(l, h)} = \mathbf{D}^{(l, h)} - \mathbf{A}^{(l, h)},
\end{equation}
 where $\mathbf{D}^{(l, h)}$ is a diagonal degree matrix. Since the attention map defines a directed graph, we distinguish between the \textit{in-degree} and \textit{out-degree} matrices. The \textit{in-degree} is computed as the sum of attention scores from preceding tokens, and due to the softmax normalization, it is uniformly 1. Therefore, we define $\mathbf{D}^{(l, h)}$ as the \textit{out-degree} matrix, which quantifies the total attention a token receives from tokens that follow it. To ensure these values remain independent of the sequence length, we normalize them by the number of subsequent tokens (i.e., the number of outgoing edges).

\begin{equation}
\label{eq:laplacian_degree}
     d^{(l,h)}_{ii} = \frac{\sum_{u}{a^{(l, h)}_{ui}}}{T-i},
\end{equation}
where $i, u \in \{0, \dots, (T-1)\}$ denote token indices. The Laplacian defined this way is bounded, i.e., $\mathbf{L}^{(l, h)}_{ij} \in \left[-1, 1\right]$ (see Appendix \ref{sec:appendix_laplacian_bounds} for proofs). Intuitively, the resulting Laplacian for each processed token represents the average attention score to previous tokens reduced by the attention score to itself. As eigenvalues of the Laplacian can summarize information flow in a graph \citep{von_luxburg_tutorial_2007, topping_understanding_2022}, we take eigenvalues of $\mathbf{L}^{(l, h)}$, which are diagonal entries due to the lower triangularity of the Laplacian matrix, and sort them:
\begin{equation}
    \Tilde{z}^{(l, h)} = \sort{\diag{\mathbf{L}^{(l, h)}}}
\end{equation}
Recently, \citep{zhu_pollmgraph_2024} found features from the entire token sequence, rather than a single token, improving hallucination detection. Similarly, \citep{kim_detecting_2024} demonstrated that information from all layers, instead of a single one in isolation, yields better results on this task. Motivated by these findings, our method uses features from all tokens and all layers as input to the probe.
Therefore, we take the top-$k$ largest values from each head and layer and concatenate them into a single feature vector $z$, where $k$ is a hyperparameter of our method:
\begin{equation}
    z = \concat_{\forall l \in L, \forall h \in H} \left[\Tilde{z}^{(l, h)}_{T}, \Tilde{z}^{(l, h)}_{T-1}, \dotsc, \Tilde{z}^{(l, h)}_{T-k}\right]
\end{equation}
Since LLMs contain dozens of layers and heads, the probe input vector $z \in \mathbb{R}^{L\cdot H\cdot k}$ can still be high-dimensional. Thus, we project it to a lower dimensionality using PCA \citep{jolliffe_principal_2016}. We call our approach $\lapeig$.
 
\section{Experimental setup}
The overview of the methodology used in this work is presented in Figure~\ref{fig:methodology}. Next, we describe each step of the pipeline in detail.

\subsection{Dataset construction}
\label{sec:datasets}
We use annotated QA datasets to construct the hallucination detection datasets and label incorrect LLM answers as hallucinations. To assess the correctness of generated answers, we followed prior work \citep{orgad_llms_2025} and adopted the \llmjudge\ approach \citep{zheng_judging_2023}, with the exception of one dataset where exact match evaluation against ground-truth answers was possible. For \llmjudge, we prompted a large LLM to classify each response as either \textit{hallucination}, \textit{non-hallucination}, or \textit{rejected}, where \textit{rejected} indicates that it was unclear whether the answer was correct, e.g., the model refused to answer due to insufficient knowledge. Based on the manual qualitative inspection of several LLMs, we employed \gptmini\ \citep{openai_gpt-4_2024} as the judge model since it provides the best trade-off between accuracy and cost. To confirm the reliability of the labels, we additionally verified agreement with the larger model, \gptnew, on $\llamabig$ and found that the agreement between models falls within the acceptable range widely adopted in the literature (see Appendix \ref{sec:appendix_cohen_kappa}).

For experiments, we selected 7 QA datasets previously utilized in the context of hallucination detection \citep{chen_inside_2024, kossen_semantic_2024, chuang_dola_2024, mitra_factlens_2024}. Specifically, we used the validation set of NQ-Open \citep{kwiatkowski_natural_2019}, comprising $3{,}610$ question-answer pairs, and the validation set of TriviaQA \citep{joshi_triviaqa_2017}, containing $7{,}983$ pairs. To evaluate our method on longer inputs, we employed the development set of CoQA \citep{reddy_coqa_2019} and the \textit{rc.nocontext} portion of the SQuADv2 \citep{rajpurkar_know_2018} datasets, with $5{,}928$ and $9{,}960$ examples, respectively. Additionally, we incorporated the QA part of the HaluEvalQA \citep{li_halueval_2023} dataset, containing $10{,}000$ examples, and the \texttt{generation} part of the TruthfulQA \citep{lin_truthfulqa_2022} benchmark with $817$ examples. Finally, we used test split of GSM8k dataset \cite{cobbe2021gsm8k} with $1{,}319$ grade school math problems, evaluated by exact match against labels. For TriviaQA, CoQA, and SQuADv2, we followed the same preprocessing procedure as \citep{chen_inside_2024}.

We generate answers using 5 open-source LLMs: \llamabig\footnote{\href{https://hf.co/meta-llama/Llama-3.1-8B-Instruct}{hf.co/meta-llama/Llama-3.1-8B-Instruct}} and \llamasmall\footnote{\href{https://huggingface.co/meta-llama/Llama-3.2-3B-Instruct}{hf.co/meta-llama/Llama-3.2-3B-Instruct}} \citep{grattafiori_llama_2024}, \philllm\footnote{\href{https://huggingface.co/microsoft/Phi-3.5-mini-instruct}{hf.co/microsoft/Phi-3.5-mini-instruct}} \citep{abdin_phi-3_2024}, \mistralnemo\footnote{\href{https://huggingface.co/mistralai/Mistral-Nemo-Instruct-2407}{hf.co/mistralai/Mistral-Nemo-Instruct-2407}} \citep{Mistral-Nemo-Instruct-2407}, \mistralsmall\footnote{\href{https://huggingface.co/mistralai/Mistral-Small-24B-Instruct-2501}{hf.co/mistralai/Mistral-Small-24B-Instruct-2501}} \citep{Mistral-Small-24B-Instruct-2501}. We use two \texttt{softmax} temperatures for each LLM when decoding ($temp \in \{0.1, 1.0\}$) and one prompt (for all datasets we used a prompt in Listing~\ref{lst:p3}, except for GSM8K in Listing~\ref{lst:gsm8k}). Overall, we evaluated hallucination detection probes on 10 LLM configurations and 7 QA datasets. We present the frequency of classes for answers from each configuration in Figure~\ref{fig:ds_sizes} (Appendix~\ref{sec:ds_sizes}).

\subsection{Hallucination Probe}
As a hallucination probe, we take a logistic regression model, using the implementation from scikit-learn \citep{pedregosa_scikit-learn_2011} with all parameters default, except for ${max\_iter{=}2000}$ and ${class\_weight{=}{''balanced''}}$. For top-$k$ eigenvalues, we tested 5 values of $k \in \{5, 10, 20, 50, 100\}$\footnote{For datasets with examples having less than 100 tokens, we stop at $k{=}50$} and selected the result with the highest efficacy. All eigenvalues are projected with PCA onto 512 dimensions, except in \textit{per-layer} experiments where there may be fewer than 512 features. In these cases, we apply PCA projection to match the input feature dimensionality, i.e., decorrelating them. As an evaluation metric, we use AUROC on the test split (additional results presenting Precision and Recall are reported in Appendix \ref{sec:appendix_extended_precision_recall}).

\subsection{Baselines}
Our method is a supervised approach for detecting hallucinations using only attention maps. For a fair comparison, we adapt the unsupervised $\attnscore$ \citep{sriramanan_llm-check_2024} by using log-determinants of each head's attention scores as features instead of summing them, and we also include the original $\attnscore$, computed as the sum of log-determinants over heads, for reference. To evaluate the effectiveness of our proposed Laplacian eigenvalues, we compare them to the eigenvalues of raw attention maps, denoted as $\attneig$. Extended results for each approach on a per-layer basis are provided in Appendix~\ref{sec:appendix_extended_method_comparison}, while Appendix~\ref{sec:appendix_hidden_states} presents a comparison with a method based on hidden states. Implementation and hardware details are provided in Appendix~\ref{sec:appendix_implementation_details}.

\section{Results}
\label{sec:results}

\begin{table*}[!htb]
    \centering
    \caption{Test AUROC for $\lapeig$ and several baseline methods. AUROC values were obtained in a single run of logistic regression training on features from a dataset generated with $temp{=}1.0$. We mark results for $\attnscore$ in \textcolor{gray}{gray} as it is an unsupervised approach, not directly comparable to the others. In \textbf{bold}, we highlight the best performance individually for each dataset and LLM. See Appendix~\ref{sec:appendix_detailed_results} for extended results.}
    \label{tab:main_results}
    \small
    \resizebox{\textwidth}{!}{%
    \begin{tabular}{llrrrrrrr}
\toprule
LLM & Feature & \multicolumn{7}{c}{Test AUROC ($\uparrow$)} \\
\cmidrule(lr){3-9}
& & CoQA & GSM8K & HaluevalQA & NQOpen & SQuADv2 & TriviaQA & TruthfulQA \\
\midrule
\textcolor{gray}{Llama3.1-8B} & \textcolor{gray}{$\attnscore$} & \textcolor{gray}{0.493} & \textcolor{gray}{0.720} & \textcolor{gray}{0.589} & \textcolor{gray}{0.556} & \textcolor{gray}{0.538} & \textcolor{gray}{0.532} & \textcolor{gray}{0.541} \\
Llama3.1-8B & $\attnlogdet$ & 0.769 & 0.826 & 0.827 & 0.793 & 0.748 & 0.842 & 0.814 \\
Llama3.1-8B & $\attneig$ & 0.782 & 0.838 & 0.819 & 0.790 & 0.768 & 0.843 & \textbf{0.833} \\
Llama3.1-8B & $\lapeig$ & \textbf{0.830} & \textbf{0.872} & \textbf{0.874} & \textbf{0.827} & \textbf{0.791} &\textbf{0.889} & 0.829 \\
\midrule
\textcolor{gray}{Llama3.2-3B} & \textcolor{gray}{$\attnscore$} & \textcolor{gray}{0.509} &  \textcolor{gray}{0.717} & \textcolor{gray}{0.588} & \textcolor{gray}{0.546} & \textcolor{gray}{0.530} & \textcolor{gray}{0.515} & \textcolor{gray}{0.581} \\
Llama3.2-3B & $\attnlogdet$ & 0.700 & 0.851  & 0.801 & 0.690 & 0.734 & 0.789 & \textbf{0.795} \\
Llama3.2-3B & $\attneig$ & 0.724 & 0.768 & 0.819 & \textbf{0.694} & 0.749 & 0.804 & 0.723 \\
Llama3.2-3B & $\lapeig$ & \textbf{0.812} & \textbf{0.870} & \textbf{0.828} & 0.693 & \textbf{0.757} & \textbf{0.832} & 0.787 \\
\midrule
\textcolor{gray}{Phi3.5} & \textcolor{gray}{$\attnscore$} & \textcolor{gray}{0.520} & \textcolor{gray}{0.666} & \textcolor{gray}{0.541} & \textcolor{gray}{0.594} & \textcolor{gray}{0.504} & \textcolor{gray}{0.540} & \textcolor{gray}{0.554} \\
Phi3.5 & $\attnlogdet$ & 0.745 & 0.842 & 0.818 & 0.815 & 0.769 & 0.848 & 0.755 \\
Phi3.5 & $\attneig$ & 0.771 & 0.794 & 0.829 & 0.798 & 0.782 & 0.850 & \textbf{0.802} \\
Phi3.5 & $\lapeig$ & \textbf{0.821} & \textbf{0.885} & \textbf{0.836} & \textbf{0.826} &\textbf{ 0.795} & \textbf{0.872} & 0.777 \\
\midrule
\textcolor{gray}{Mistral-Nemo} & \textcolor{gray}{$\attnscore$} & \textcolor{gray}{0.493} & \textcolor{gray}{0.630} & \textcolor{gray}{0.531} & \textcolor{gray}{0.529} & \textcolor{gray}{0.510} & \textcolor{gray}{0.532} & \textcolor{gray}{0.494} \\
Mistral-Nemo & $\attnlogdet$ & 0.728 & 0.856 & 0.798 & 0.769 & 0.772 & 0.812 & \textbf{0.852} \\
Mistral-Nemo & $\attneig$ & 0.778 & 0.842 & 0.781 & 0.761 & 0.758 & 0.821 & 0.802 \\
Mistral-Nemo & $\lapeig$ & \textbf{0.835} & \textbf{0.890} & \textbf{0.833} & \textbf{0.795} & \textbf{0.812}& \textbf{0.865} & 0.828 \\
\midrule
\textcolor{gray}{Mistral-Small-24B} & \textcolor{gray}{$\attnscore$} & \textcolor{gray}{0.516} & \textcolor{gray}{0.576} & \textcolor{gray}{0.504} & \textcolor{gray}{0.462} & \textcolor{gray}{0.455} & \textcolor{gray}{0.463} & \textcolor{gray}{0.451} \\
Mistral-Small-24B & $\attnlogdet$ & 0.766 & 0.853 & 0.842 & 0.747 & 0.753 & 0.833 & 0.735 \\
Mistral-Small-24B & $\attneig$ & 0.805  & 0.856 & 0.848 & 0.751 & 0.760 & 0.844 & \textbf{0.765} \\
Mistral-Small-24B & $\lapeig$ & \textbf{0.861} & \textbf{0.925} & \textbf{0.882} & \textbf{0.791} & \textbf{0.820} & \textbf{0.876} & 0.748 \\
\bottomrule
\end{tabular}

    }
\end{table*}

Table~\ref{tab:main_results} presents the results of our method compared to the baselines. $\lapeig$ achieved the best performance among all tested methods on 6 out of 7 datasets. Moreover, our method consistently performs well across all 5 LLM architectures ranging from 3 up to 24 billion parameters. TruthfulQA was the only exception where $\lapeig$ was the second-best approach, yet it might stem from the small size of the dataset or severe class imbalance (depicted in Figure~\ref{fig:ds_sizes}). In contrast, using eigenvalues of vanilla attention maps in $\attneig$ leads to worse performance, which suggests that transformation to Laplacian is the crucial step to uncover latent features of an LLM corresponding to hallucinations. In Appendix~\ref{sec:appendix_detailed_results}, we show that $\lapeig$ consistently demonstrates a smaller generalization gap, i.e., the difference between training and test performance is smaller for our method. While the $\attnscore$ method performed poorly, it is fully unsupervised and should not be directly compared to other approaches. However, its supervised counterpart -- $\attnlogdet$ -- remains inferior to methods based on spectral features, namely $\attneig$ and $\lapeig$. In Table~\ref{tab:detailed_results} in Appendix~\ref{sec:appendix_extended_method_comparison}, we present extended results, including \textit{per-layer} and \textit{all-layers} breakdowns, two temperatures used during answers generation, and a comparison between training and test AUROC. Moreover, compared to probes based on hidden states, our method performs best in most of the tested settings, as shown in Appendix \ref{sec:appendix_hidden_states}.

\section{Ablation studies}
To better understand the behavior of our method under different conditions, we conduct a comprehensive ablation study. This analysis provides valuable insights into the factors driving the $\lapeig$ performance and highlights the robustness of our approach across various scenarios. In order to ensure reliable results, we perform all studies on the TriviaQA dataset, which has a moderate input size and number of examples.

\subsection{How does the number of eigenvalues influence performance?}
\label{sec:ablation_topk}
First, we verify how the number of eigenvalues influences the performance of the hallucination probe and present results for $\mistralsmall$ in Figure~\ref{fig:top_k_ablation} (results for all models are showcased in Figure \ref{fig:appendix_top_k_ablation} in Appendix \ref{sec:appendix_detailed_ablation}). Generally, using more eigenvalues improves performance, but there is less variation in performance among different values of $k$ for $\lapeig$ compared to the baseline. Moreover, $\lapeig$ achieves significantly better performance with smaller input sizes, as $\attneig$ with the largest $k{=}100$ fails to surpass $\lapeig$'s performance at $k{=}5$. These results confirm that spectral features derived from the Laplacian carry a robust signal indicating the presence of hallucinations and highlight the strength of our method.

\begin{figure}[!htb]
    \centering
    \includegraphics[width=\columnwidth]{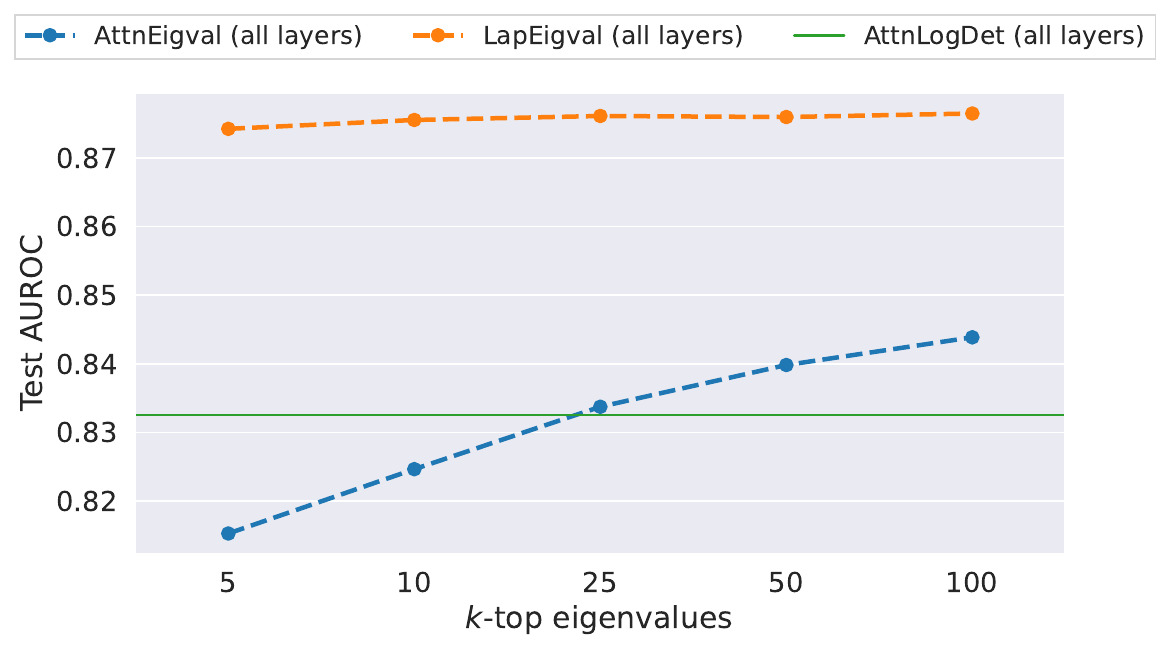}
    \caption{Probe performance across different top-$k$ eigenvalues: $k \in \{5, 10, 25, 50, 100\}$ for TriviaQA dataset with $temp{=}1.0$ and $\mistralsmall$ LLM.}
    \label{fig:top_k_ablation}
\end{figure}

\subsection{Does using all layers at once improve performance?}
\label{sec:ablation_layer}
\begin{figure}[!htb]
    \centering
    \includegraphics[width=\columnwidth]{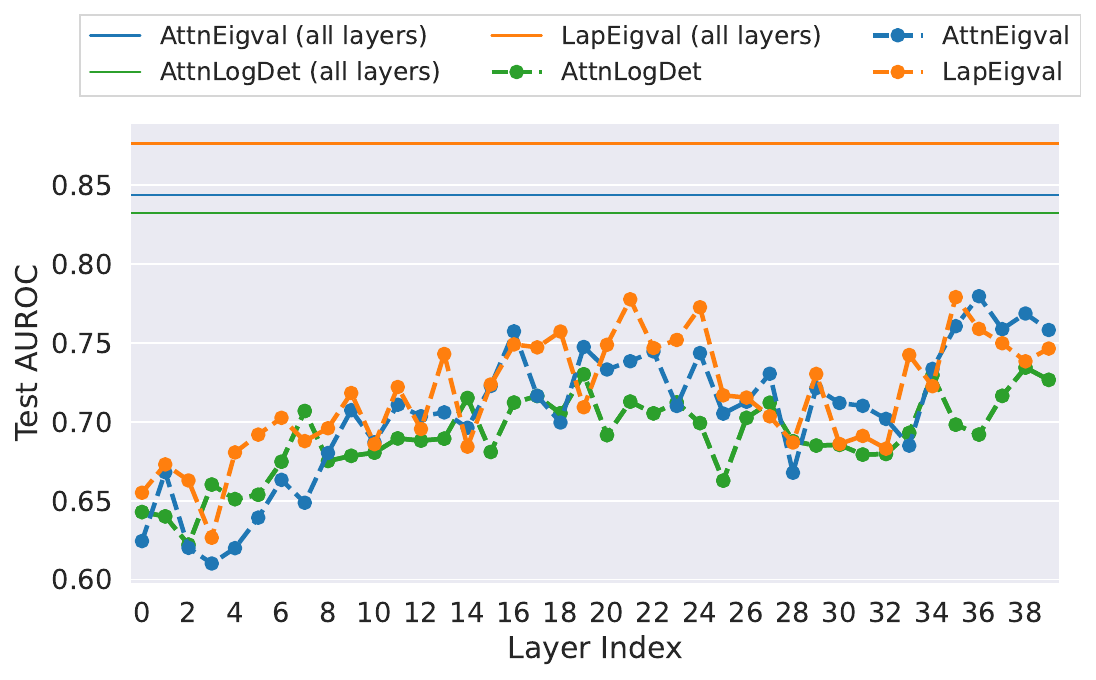}
    \caption{Analysis of model performance across different layers for $\mistralsmall$ and TriviaQA dataset with $temp{=}1.0$ and $k{=}100$ top eigenvalues (results for models operating on all layers provided for reference).}
    \label{fig:layer_idx_ablation}
\end{figure}

Second, we demonstrate that using all layers of an LLM instead of a single one improves performance. In Figure~\ref{fig:layer_idx_ablation}, we compare \textit{per-layer} to \textit{all-layer} efficacy for $\mistralsmall$ (results for all models are showcased in Figure \ref{fig:appendix_layer_idx_ablation} in Appendix \ref{sec:appendix_detailed_ablation}). For the \textit{per-layer} approach, better performance is generally achieved with deeper LLM layers. Notably, peak performance varies across LLMs, requiring an additional search for each new LLM. In contrast, the \textit{all-layer} probes consistently outperform the best \textit{per-layer} probes across all LLMs. This finding suggests that information indicating hallucinations is spread across many layers of LLM, and considering them in isolation limits detection accuracy. Further, Table~\ref{tab:detailed_results} in Appendix~\ref{sec:appendix_detailed_results} summarises outcomes for the two variants on all datasets and LLM configurations examined in this work.

\label{sec:all_layers_vs_per_layer}

\subsection{Does sampling temperature influence results?}
\label{sec:ablation_temperature}

\begin{figure}[!htb]
    \centering
    \includegraphics[width=\columnwidth]{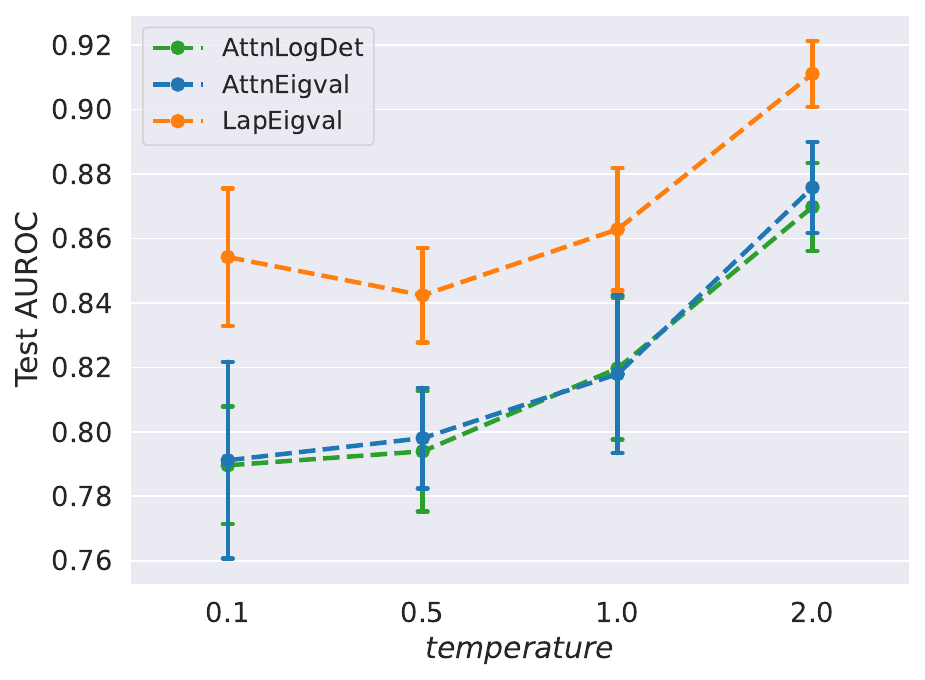}
    \caption{Test AUROC for different sampling $temp$ values during answer decoding on the TriviaQA dataset, using $k{=}100$ eigenvalues for $\lapeig$ and $\attneig$ with the \llamabig\ LLM. Error bars indicate the standard deviation over 10 balanced samples containing $N=1000$ examples per class.}
    \label{fig:temperature_ablation}
\end{figure}

Here, we compare $\lapeig$ to the baselines on hallucination datasets, where each dataset contains answers generated at a specific decoding temperature. Higher temperatures typically produce more hallucinated examples \citep{lee_mathematical_2023, renze_effect_2024}, leading to dataset imbalance. Thus, to mitigate the effect of data imbalance, we sample a subset of $1{,}000$ hallucinated and $1{,}000$ non-hallucinated examples $10$ times for each temperature and train hallucination probes. Interestingly, in Figure~\ref{fig:temperature_ablation}, we observe that all models improve their performance at higher temperatures, but $\lapeig$ consistently achieves the best accuracy on all considered temperature values. The correlation of efficacy with temperature may be attributed to differences in the characteristics of hallucinations at higher temperatures compared to lower ones \citep{renze_effect_2024}. Also, hallucination detection might be facilitated at higher temperatures due to underlying properties of \texttt{softmax} function \citep{velickovic_softmax_2024}, and further exploration of this direction is left for future work.

\subsection{How does \texorpdfstring{$\lapeig$}{\textit{LapEigvals}} generalizes?}
\label{sec:generalization}
To check whether our method generalizes across datasets, we trained the hallucination probe on features from the training split of one QA dataset and evaluated it on the features from the test split of a different QA dataset. Due to space limitations, we present results for selected datasets and provide extended results and absolute efficacy values in Appendix~\ref{sec:appendix_generalization}. Figure \ref{fig:generalization_main} showcases the percent drop in Test AUROC when using a different training dataset compared to training and testing on the same QA dataset. We can observe that $\lapeig$ provides a performance drop comparable to other baselines, and in several cases, it generalizes best. Interestingly, all methods exhibit poor generalization on TruthfulQA and GSM8K. We hypothesize that the weak performance on TruthfulQA arises from its limited size and class imbalance, whereas the difficulty on GSM8K likely reflects its distinct domain, which has been shown to hinder hallucination detection \citep{orgad_llms_2025}. Additionally, in Appendix~\ref{sec:appendix_generalization}, we show that $\lapeig$ achieves the highest test performance in all scenarios (except for TruthfulQA).
\begin{figure*}[!htb]
    \centering
    \includegraphics[width=\textwidth]{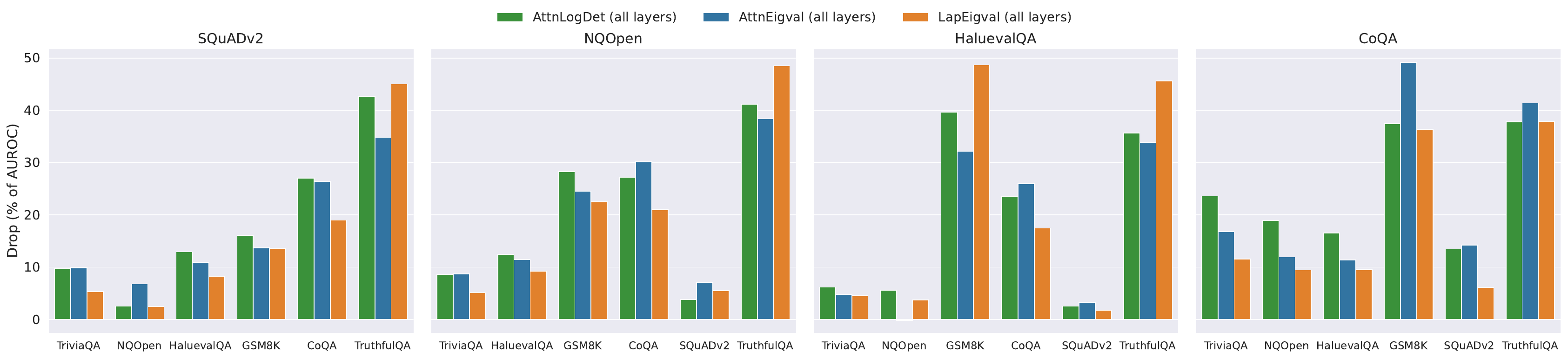}
    \caption{Generalization across datasets measured as a percent performance drop in Test AUROC (less is better) when trained on one dataset and tested on the other. Training datasets are indicated in the plot titles, while test datasets are shown on the $x$-axis. Results computed on \llamabig\ with $k{=}100$ top eigenvalues and $temp{=}1.0$. Results for all datasets are presented in Appendix~\ref{sec:appendix_generalization}.}
    \label{fig:generalization_main}
\end{figure*}

\subsection{How does performance vary across prompts?}
\label{sec:ablation_prompts}
Lastly, to assess the stability of our method across different prompts used for answer generation, we compared the results of the hallucination probes trained on features regarding four distinct prompts, the content of which is included in Appendix~\ref{sec:appendix_prompts}. As shown in Table~\ref{tab:prompt_comparison}, $\lapeig$ consistently outperforms all baselines across all four prompts. While we can observe variations in performance across prompts, $\lapeig$ demonstrates the lowest standard deviation ($0.05$) compared to $\attnlogdet$ ($0.016$) and $\attneig$ ($0.07$), indicating its greater robustness.

\begin{table}[!htb]
    \centering
    \caption{Test AUROC across four different prompts for answers on the TriviaQA dataset using \llamabig\, with $temp{=}1.0$ and $k{=}50$ (some prompts have led to fewer than 100 tokens). Prompt $\boldsymbol{p_3}$ was the main one used to compare our method to baselines, as presented in Tables~\ref{tab:main_results}.}
    \resizebox{\columnwidth}{!}{
        \begin{tabular}{lrrrr}
\toprule
Feature & \multicolumn{4}{c}{Test AUROC ($\uparrow$)} \\
\cmidrule(lr){2-5}
& $p_1$ & $p_2$ & $\boldsymbol{p_3}$ & $p_4$ \\
\midrule
$\attnlogdet$ & 0.847 & 0.855 & 0.842 & 0.860 \\
$\attneig$ & 0.840 & 0.870 & 0.842 & 0.875 \\
$\lapeig$ & \textbf{0.882} & \textbf{0.890} & \textbf{0.888} & \textbf{0.895} \\
\bottomrule
\end{tabular}
    }
    \label{tab:prompt_comparison}
\end{table}

\section{Related Work}
Hallucinations in LLMs were proved to be inevitable \citep{xu_hallucination_2024}, and to detect them, one can leverage either \textit{black-box} or \textit{white-box} approaches. The former approach uses only the outputs from an LLM, while the latter uses hidden states, attention maps, or logits corresponding to generated tokens.

Black-box approaches focus on the text generated by LLMs. For instance, \citep{li_dawn_2024} verified the truthfulness of factual statements using external knowledge sources, though this approach relies on the availability of additional resources. Alternatively, \textit{SelfCheckGPT} \citep{manakul_selfcheckgpt_2023} generates multiple responses to the same prompt and evaluates their consistency, with low consistency indicating potential hallucination.

White-box methods have emerged as a promising approach for detecting hallucinations \citep{farquhar_detecting_2024, azaria_internal_2023, arteaga_hallucination_2024, orgad_llms_2025}. These methods are universal across all LLMs and do not require additional domain adaptation compared to black-box ones \citep{farquhar_detecting_2024}. They draw inspiration from seminal works on analyzing the internal states of simple neural networks \citep{alain_understanding_2016}, which introduced \textit{linear classifier probes} -- models operating on the internal states of neural networks. Linear probes have been widely applied to the internal states of LLMs, notably for detecting hallucinations. 

One of the first such probes was SAPLMA \citep{azaria_internal_2023}, which demonstrated that one could predict the correctness of generated text straight from LLM's hidden states. Further, the INSIDE method \citep{chen_inside_2024} tackled hallucination detection by sampling multiple responses from an LLM and evaluating consistency between their hidden states using a normalized sum of the eigenvalues from their covariance matrix. Also, \citep{farquhar_detecting_2024} proposed a complementary probabilistic approach, employing entropy to quantify the model's intrinsic uncertainty. Their method involves generating multiple responses, clustering them by semantic similarity, and calculating Semantic Entropy using an appropriate estimator. To address concerns regarding the validity of LLM probes, \citep{marks_geometry_2024} introduced a high-quality QA dataset with simple \textit{true}/\textit{false} answers and causally demonstrated that the truthfulness of such statements is linearly represented in LLMs, which supports the use of probes for short texts.

Self-consistency methods \citep{liang_internal_2024}, like INSIDE or Semantic Entropy, require multiple runs of an LLM for each input example, which substantially lowers their applicability. Motivated by this limitation, \citep{kossen_semantic_2024} proposed to use \textit{Semantic Entropy Probe}, which is a small model trained to predict expensive Semantic Entropy \citep{farquhar_detecting_2024} from LLM's hidden states. Notably, \citep{orgad_llms_2025} explored how LLMs encode information about truthfulness and hallucinations. First, they revealed that truthfulness is concentrated in specific tokens. Second, they found that probing classifiers on LLM representations do not generalize well across datasets, especially across datasets requiring different skills, which we confirmed in Section~\ref{sec:generalization}. Lastly, they showed that the probes could select the correct answer from multiple generated answers with reasonable accuracy, meaning LLMs make mistakes at the decoding stage, besides knowing the correct answer.

Recent studies have started to explore hallucination detection exclusively from attention maps. \citep{chuang_lookback_2024} introduced the \textit{lookback ratio}, which measures how much attention LLMs allocate to relevant input parts when answering questions based on the provided context. The work most closely related to ours is \citep{sriramanan_llm-check_2024}, which introduces the $\attnscore$ method. Although the process is unsupervised and computationally efficient, the authors note that its performance can depend highly on the specific layer from which the score is extracted. Compared to $\attnscore$, our method is fully supervised and grounded in graph theory, as we interpret inference in LLM as a graph. While $\attnscore$ aggregates only the attention diagonal to compute its log-determinant, we instead derive features from the graph Laplacian, which captures all attention scores (see Eq. \eqref{eq:laplacian} and \eqref{eq:laplacian_degree}). Additionally, we utilize all layers for detecting hallucination rather than a single one, demonstrating effectiveness of this approach. We also demonstrate that it performs poorly on the datasets we evaluated. Nonetheless, we drew inspiration from their approach, particularly using the lower triangular structure of matrices when constructing features for the hallucination probe.

\section{Conclusions}
In this work, we demonstrated that the spectral features of LLMs' attention maps, specifically the eigenvalues of the Laplacian matrix, carry a signal capable of detecting hallucinations. Specifically, we proposed the $\lapeig$ method, which employs the top-$k$ eigenvalues of the Laplacian as input to the hallucination detection probe. Through extensive evaluations, we empirically showed that our method consistently achieves state-of-the-art performance among all tested approaches. Furthermore, multiple ablation studies demonstrated that our method remains stable across varying numbers of eigenvalues, diverse prompts, and generation temperatures while offering reasonable generalization.

In addition, we hypothesize that self-supervised learning \citep{balestriero_cookbook_2023} could yield a more robust and generalizable approach while uncovering non-trivial intrinsic features of attention maps. Notably, results such as those in Section~\ref{sec:ablation_temperature} suggest intriguing connections to recent advancements in LLM research \citep{velickovic_softmax_2024, barbero_transformers_2024}, highlighting promising directions for future investigation.

\section*{Limitations}
\textit{\textbf{Supervised method}} In our approach, one must provide labelled hallucinated and non-hallucinated examples to train the hallucination probe. While this can be handled by the $\llmjudge$, it might introduce some noise or pose a risk of overfitting. \textit{\textbf{Limited generalization across LLM architectures}} The method is incompatible with LLMs having different head and layer configurations. Developing architecture-agnostic hallucination probes is left for future work. \textit{\textbf{Minimum length requirement}} Computing $\topk$ Laplacian eigenvalues demands attention maps of at least $k$ tokens (e.g., $k{=}100$ require 100 tokens). \textbf{\textit{Open LLMs}} Our method requires access to the internal states of LLM thus it cannot be applied to closed LLMs.
\textit{\textbf{Risks}} Please note that the proposed method was tested on selected LLMs and English data, so applying it to untested domains and tasks carries a considerable risk without additional validation.

\section*{Acknowledgements}
We sincerely thank Piotr Bielak for his valuable review and insightful feedback, which helped improve this work. This work was funded by the European Union under the Horizon Europe grant OMINO – Overcoming Multilevel INformation Overload (grant number 101086321, \url{https://ominoproject.eu/}). Views and opinions expressed are those of the authors alone and do not necessarily reflect those of the European Union or the European Research Executive Agency. Neither the European Union nor the European Research Executive Agency can be held responsible for them. It was also co-financed with funds from the Polish Ministry of Education and Science under the programme entitled International Co-Financed Projects, grant no. 573977. We gratefully acknowledge the Wroclaw Centre for Networking and Supercomputing for providing the computational resources used in this work. This work was co-funded by the National Science Centre, Poland under CHIST-ERA Open \& Re-usable Research Data \& Software  (grant number 2022/04/Y/ST6/00183). The authors used ChatGPT to improve the clarity and readability of the manuscript.


\bibliography{references}

\clearpage
\appendix
\section{Details of motivational study}
\label{sec:appendix_stat_test}
We present a detailed description of the procedure used to obtain the results presented in Section~\ref{sec:motivation}, along with additional results for other datasets and LLMs. 

Our goal was to test whether $\attnscore$ and eigenvalues of Laplacian matrix (used by our $\lapeig$) differ significantly when examples are split into hallucinated and non-hallucinated groups. To this end, we used 7 datasets (Section~\ref{sec:datasets}) and ran inference with 5 LLMs (Section~\ref{sec:datasets}) using $temp{=}0.1$. From the extracted attention maps, we computed $\attnscore$ \citep{sriramanan_llm-check_2024}, defined as the log-determinant of the attention matrices. Unlike the original work, we did not aggregate scores across heads, but instead analyzed them at the single-head level. For $\lapeig$, we constructed the Laplacian as defined in Section~\ref{sec:method}, extracted the 10 largest eigenvalues per head, and applied the same single-head analysis as for $\attneig$. Finally, we performed the Mann–Whitney U test \citep{mann1947test} using the SciPy implementation \citep{virtanen_scipy_2020} and collected the resulting $p$-values

Table~\ref{tab:full_motivation_results} presents the percentage of heads having a statistically significant difference in feature values between hallucinated and non-hallucinated examples, as indicated by $p < 0.05$ from the Mann-Whitney U test. These results show that the Laplacian eigenvalues better distinguish between the two classes for almost all considered LLMs and datasets.

\begin{table}[htb]
    \centering
    \caption{Percentage of heads having a statistically significant difference in feature values between hallucinated and non-hallucinated examples, as indicated by $p < 0.05$ from the Mann-Whitney U test. Results were obtained for $\attnscore$ and the 10 largest Laplacian eigenvalues on 6 datasets and 5 LLMs.}
    \resizebox{\columnwidth}{!}{
        \begin{tabular}{llrr}
\toprule
LLM & Dataset & \multicolumn{2}{c}{$\% \text{ of } p < 0.05$} \\
\cmidrule(lr){3-4}
& & AttnScore & Laplacian eigvals \\
\midrule
Llama3.1-8B & CoQA & 40 & 87 \\
Llama3.1-8B & GSM8K & 83 & 70 \\
Llama3.1-8B & HaluevalQA & 91 & 93 \\
Llama3.1-8B & NQOpen & 78 & 83 \\
Llama3.1-8B & SQuADv2 & 70 & 81 \\
Llama3.1-8B & TriviaQA & 80 & 91 \\
Llama3.1-8B & TruthfulQA & 40 & 60 \\
\midrule
Llama3.2-3B & CoQA & 50 & 79 \\
Llama3.2-3B & GSM8K & 74 & 67 \\
Llama3.2-3B & HaluevalQA & 91 & 93 \\
Llama3.2-3B & NQOpen & 81 & 84 \\
Llama3.2-3B & SQuADv2 & 69 & 74 \\
Llama3.2-3B & TriviaQA & 81 & 87 \\
Llama3.2-3B & TruthfulQA & 40 & 62 \\
\midrule
Phi3.5 & CoQA & 45 & 81 \\
Phi3.5 & GSM8K & 67 & 69 \\
Phi3.5 & HaluevalQA & 80 & 86 \\
Phi3.5 & NQOpen & 73 & 80 \\
Phi3.5 & SQuADv2 & 81 & 82 \\
Phi3.5 & TriviaQA & 86 & 92 \\
Phi3.5 & TruthfulQA & 41 & 53 \\
\midrule
Mistral-Nemo & CoQA & 35 & 78 \\
Mistral-Nemo & GSM8K & 90 & 71 \\
Mistral-Nemo & HaluevalQA & 78 & 82 \\
Mistral-Nemo & NQOpen & 64 & 57 \\
Mistral-Nemo & SQuADv2 & 54 & 56 \\
Mistral-Nemo & TriviaQA & 71 & 74 \\
Mistral-Nemo & TruthfulQA & 40 & 50 \\
\midrule
Mistral-Small-24B & CoQA & 28 & 78 \\
Mistral-Small-34B & GSM8K & 75 & 72 \\
Mistral-Small-24B & HaluevalQA & 68 & 70 \\
Mistral-Small-24B & NQOpen & 45 & 51 \\
Mistral-Small-24B & SQuADv2 & 75 & 82 \\
Mistral-Small-24B & TriviaQA & 65 & 70 \\
Mistral-Small-24B & TruthfulQA & 43 & 52 \\
\bottomrule
\end{tabular}
    }
    \label{tab:full_motivation_results}
\end{table}

\section{Bounds of the Laplacian}
\label{sec:appendix_laplacian_bounds}
In the following section, we prove that the Laplacian defined in \ref{sec:method} is bounded and has at least one zero eigenvalue. We denote eigenvalues as $\lambda_{i}$, and provide derivation for a single layer and head, which holds also after stacking them together into a single graph (set of per-layer graphs). For clarity, we omit the superscript ${(l, h)}$ indicating layer and head.

\begin{lemma}
    The Laplacian eigenvalues are bounded: $-1 \leq \lambda_i \leq 1$.
\end{lemma}
\begin{proof}
    Due to the lower-triangular structure of the Laplacian, its eigenvalues lie on the diagonal and are given by:
    \begin{equation*}
        \lambda_i = \mathbf{L}_{ii} = d_{ii} - a_{ii}
    \end{equation*}
    The out-degree is defined as: 
    \begin{equation*}
        d_{ii} = \frac{\sum_{u}{a_{ui}}}{T-i},
    \end{equation*}
    Since $0 \leq a_{ui} \leq 1$, the sum in the numerator is upper bounded by $T-i$, therefore $d_{ii} \leq 1$, and consequently $\lambda_i = \mathbf{L}_{ii} \leq 1$, which concludes upper-bound part of the proof.
    
    Recall that eigenvalues lie on the main diagonal of the Laplacian, hence  
    $\lambda_i = \frac{\sum_{u}{a_{uj}}}{T-i} - a_{ii}$. To find the lower bound of $\lambda_i$, we need to minimize $X = \frac{\sum_{u}{a_{uj}}}{T-i}$ and maximize $Y = a_{ii}$. 
    First, we note that $X$'s denominator is always positive $T - i > 0$, since $i \in \{0 \dots(T-1)\}$ (as defined by Eq. \eqref{eq:laplacian_degree}). For the numerator, we recall that $0 \leq a_{ui} \leq 1$; therefore, the sum has its minimum at 0, hence $X \geq 0$. Second, to maximize $Y = a_{ii}$, we can take maximum of  $0 \leq a_{ii} \leq 1$ which is $1$. Finally, $X - Y = -1$, consequently $\mathbf{L}_{ii} \geq -1$, which concludes the lower-bound part of the proof.
\end{proof}

\begin{lemma}
    For every $\mathbf{L}_{ii}$, there exists at least one zero-eigenvalue, and it corresponds to the last token $T$, i.e., $\lambda_T = 0$.
\end{lemma}
\begin{proof}
Recall that eigenvalues lie on the main diagonal of the Laplacian, hence $\lambda_i = \frac{\sum_{u}{a_{uj}}}{T-i} - a_{ii}$. Consider last token, wherein the sum in the numerator reduces to $\sum_{u}{a_{uj}} = a_{TT}$, denominator becomes $T-i = T - (T-1) = 1$, thus $\lambda_T = \frac{a_{TT}}{1} - a_{TT} = 0$.
\end{proof}

\section{Implementation details}
\label{sec:appendix_implementation_details}
In our experiments, we used HuggingFace Transformers \citep{wolf_transformers_2020}, PyTorch \citep{ansel_pytorch_2024}, and scikit-learn \citep{pedregosa_scikit-learn_2011}. We utilized Pandas \citep{team_pandas-devpandas_2020} and Seaborn \citep{waskom_seaborn_2021} for visualizations and analysis. To version data, we employed DVC \citep{ruslan_kuprieiev_dvc_2025}. The Cursor IDE was employed to assist with code development. We performed LLM inference and acquired attention maps using a single Nvidia A40 with 40GB VRAM, except for $\mistralsmall$ for which we used Nvidia H100 with 96GB VRAM. Training the hallucination probe was done using the CPU only. To compute labels using the \llmjudge\ approach, we leveraged \gptmini\ model available through OpenAI API. Detailed hyperparameter settings and code to reproduce the experiments are available in the public Github repository: \url{https://github.com/graphml-lab-pwr/lapeigvals}.

\section{Details of QA datasets}
\begin{figure*}[htbp]
    \centering
    \begin{subfigure}{0.48\textwidth}
        \includegraphics[width=\textwidth]{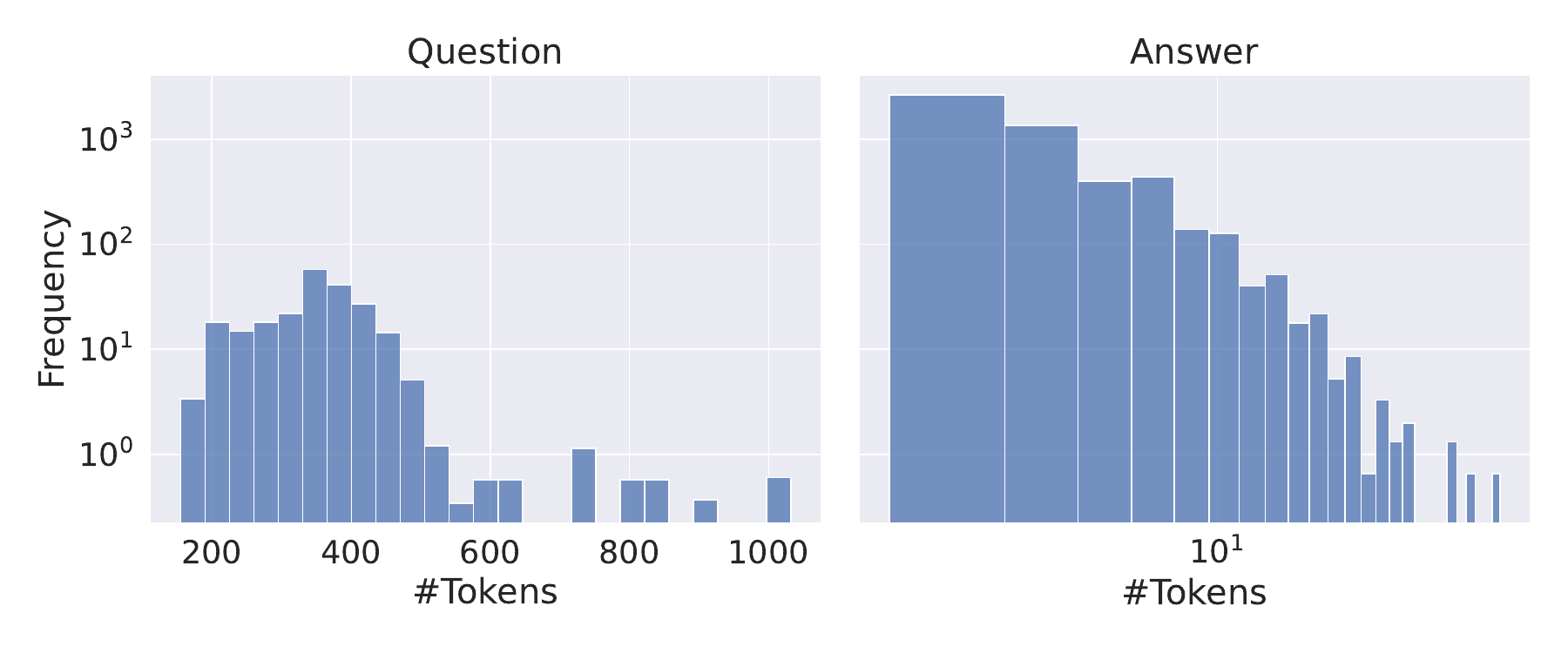}
        \caption{CoQA}
        \label{fig:token_coqa}
    \end{subfigure}
    \hfill
    \begin{subfigure}{0.48\textwidth}
        \includegraphics[width=\textwidth]{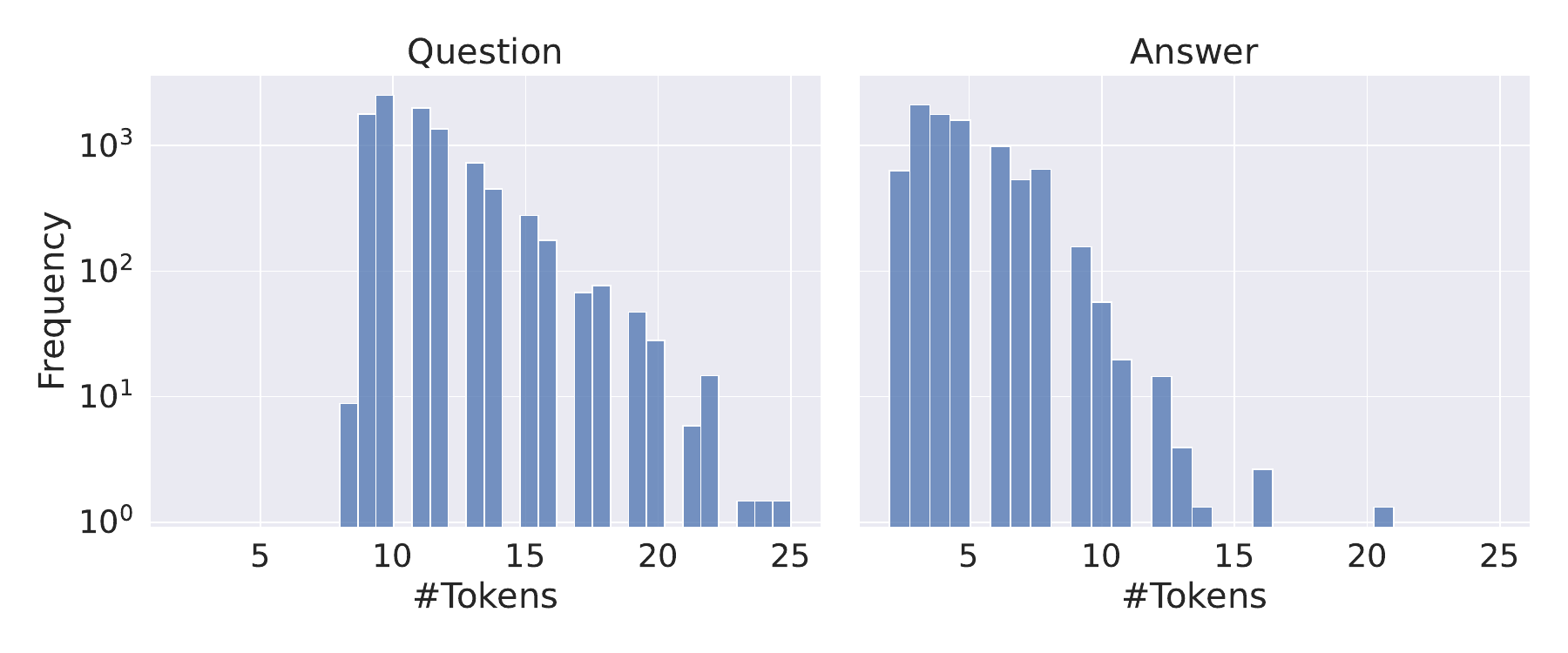}
        \caption{NQ-Open}
        \label{fig:tokens_nq_open}
    \end{subfigure}

    \vspace{1em}
    
    \begin{subfigure}{0.48\textwidth}
        \includegraphics[width=\textwidth]{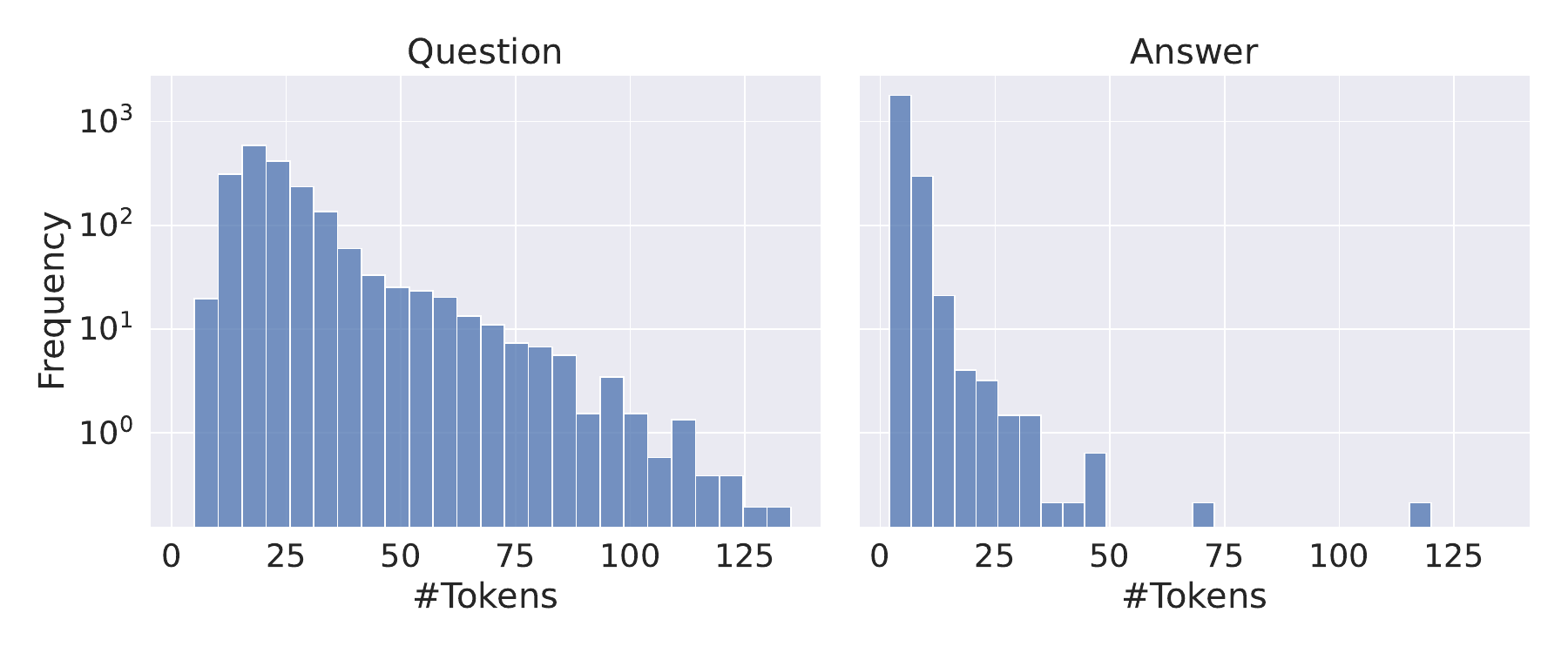}
        \caption{HaluEvalQA}
        \label{fig:tokens_halueval_qa}
    \end{subfigure}
    \hfill
    \begin{subfigure}{0.48\textwidth}
        \includegraphics[width=\textwidth]{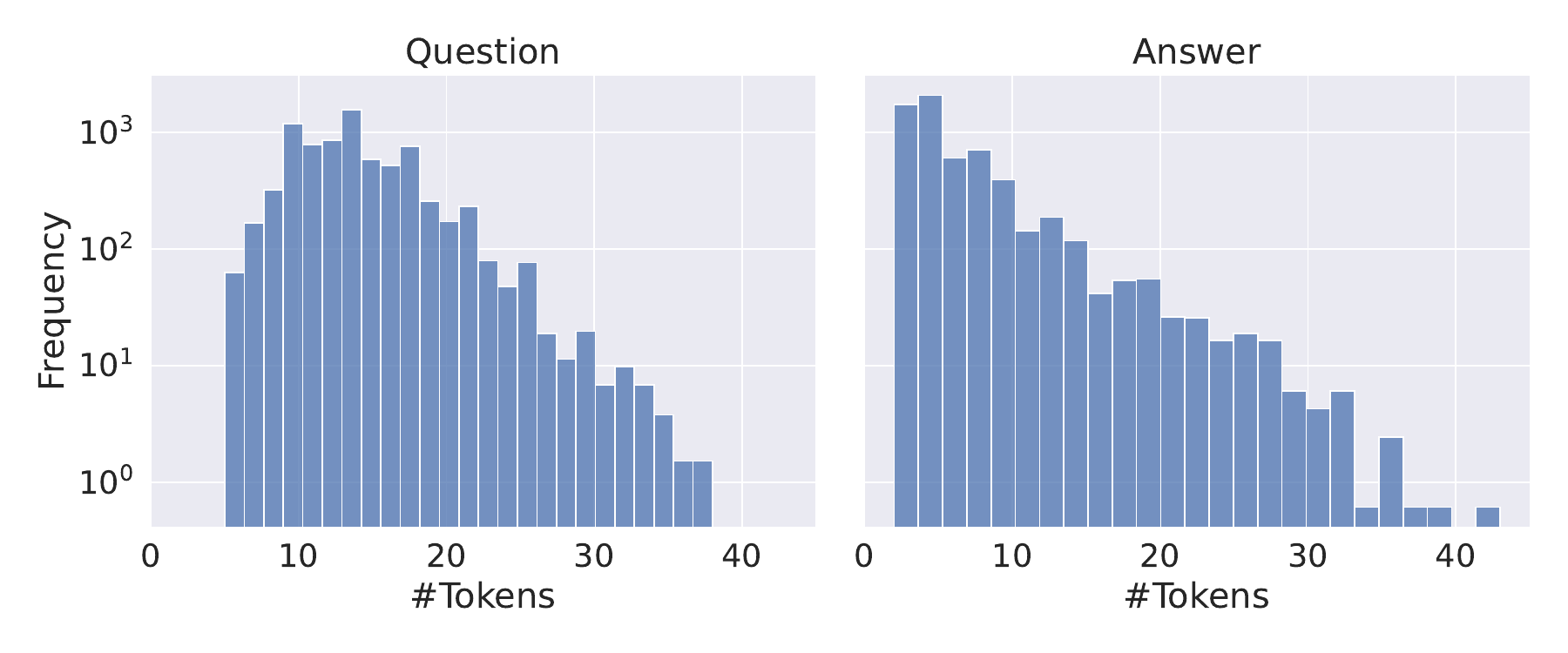}
        \caption{SQuADv2}
        \label{fig:tokens_squad_v2}
    \end{subfigure}

    \vspace{1em}

    \begin{subfigure}{0.48\textwidth}
        \includegraphics[width=\textwidth]{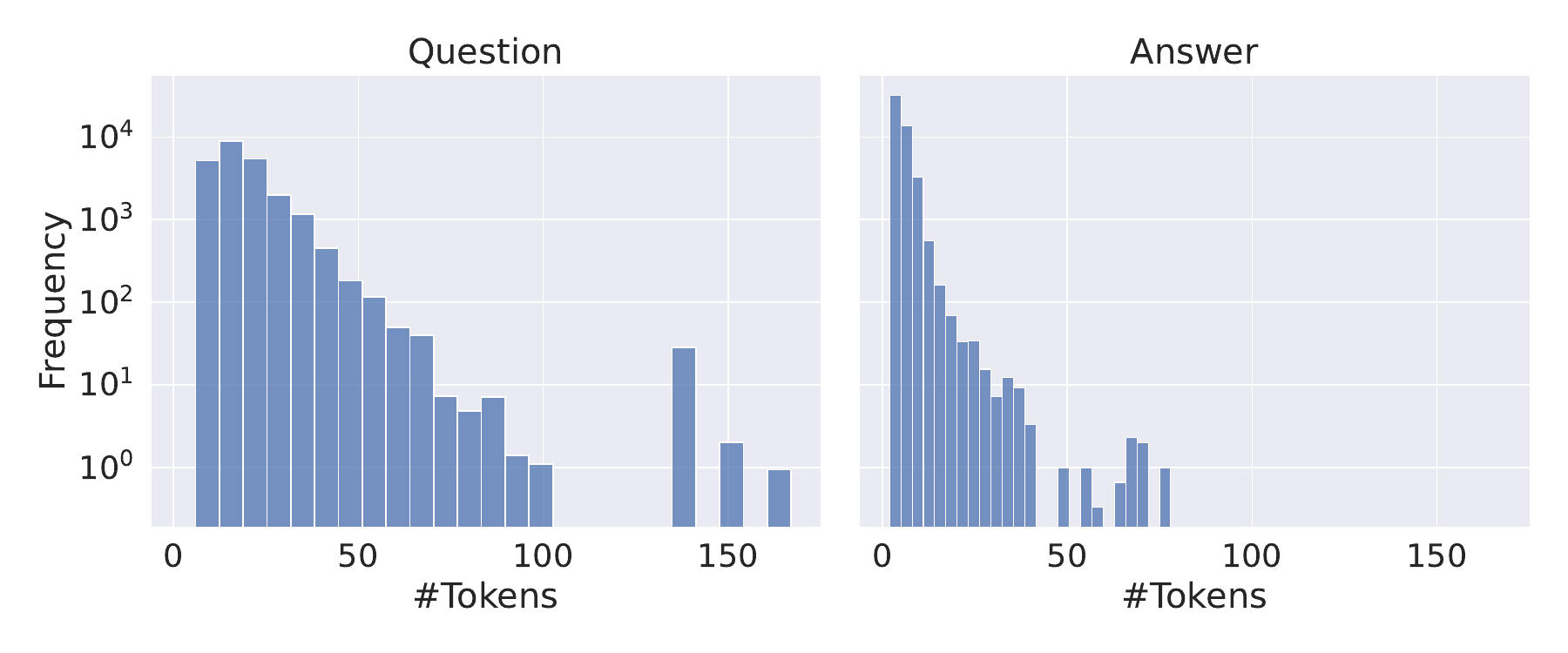}
        \caption{TriviaQA}
        \label{fig:tokens_trivia_qa}
    \end{subfigure}
    \hfill
    \begin{subfigure}{0.48\textwidth}
        \includegraphics[width=\textwidth]{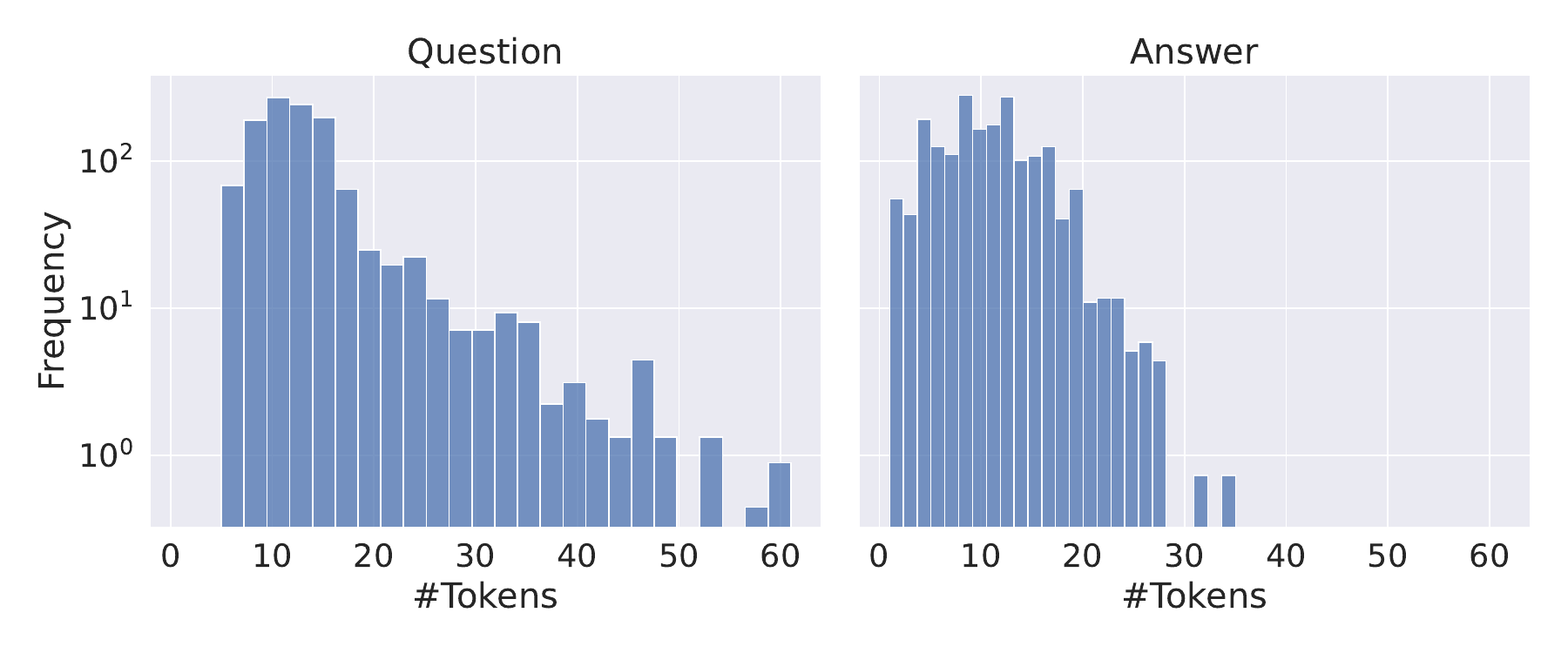}
        \caption{TruthfulQA}
        \label{fig:tokens_truthful_qa}
    \end{subfigure}

    \vspace{1em}

    \begin{subfigure}{0.48\textwidth}
        \includegraphics[width=\textwidth]{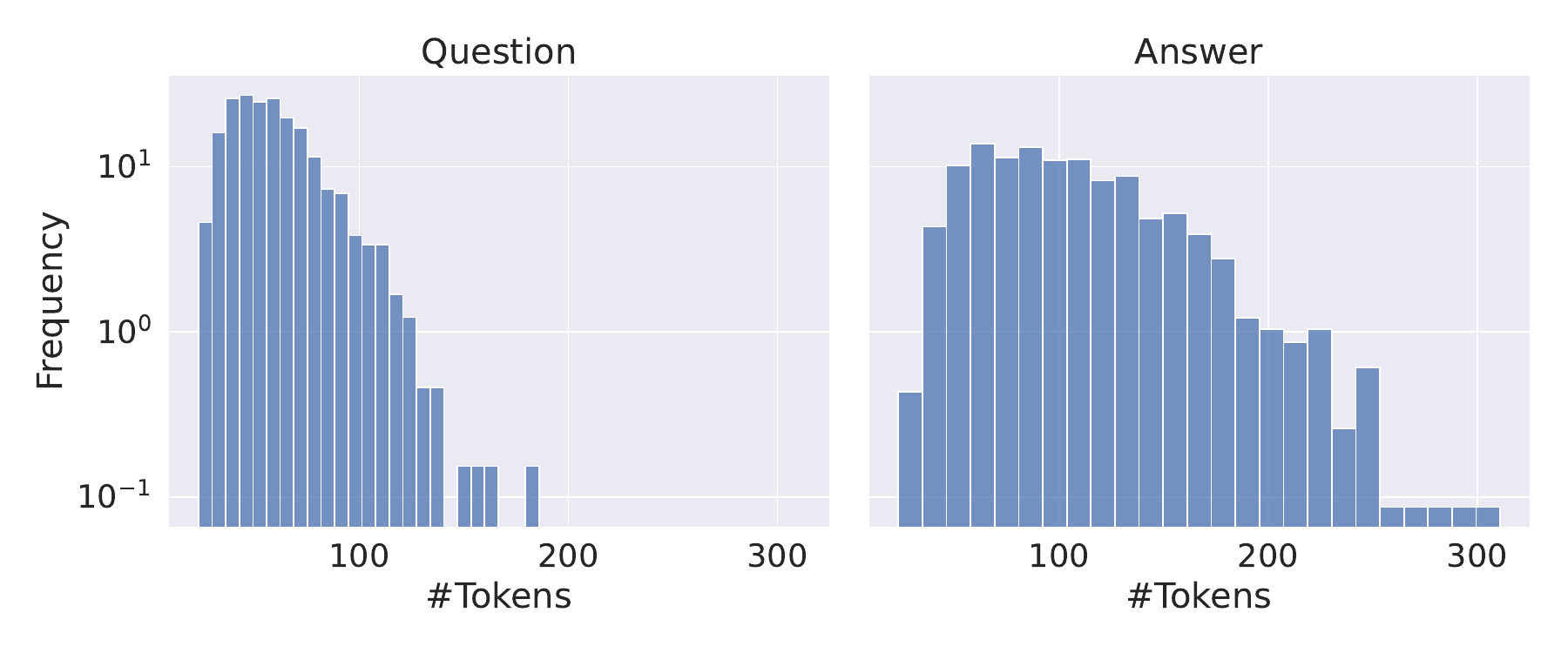}
        \caption{GSM8K}
        \label{fig:tokens_gsm8k}
    \end{subfigure}

    \caption{Token count histograms for the datasets used in our experiments. Token counts were computed separately for each example's $question$ (left) and gold $answer$ (right) using the \texttt{meta-llama/Llama-3.1-8B-Instruct} tokenizer. In cases with multiple answers, they were flattened into one.}
    \label{fig:token_stats}
\end{figure*}

We used 7 open and publicly available question answering datasets: NQ-Open \citep{kwiatkowski_natural_2019} (CC-BY-SA-3.0 license), SQuADv2 \citep{rajpurkar_know_2018} (CC-BY-SA-4.0 license), TruthfulQA (Apache-2.0 license) \citep{lin_truthfulqa_2022}, HaluEvalQA (MIT license) \citep{li_halueval_2023}, CoQA \citep{reddy_coqa_2019} (domain-dependent licensing, detailed on \url{https://stanfordnlp.github.io/coqa/}), TriviaQA (Apache 2.0 license), GSM8K \citep{cobbe2021gsm8k}(MIT license). Research purposes fall into the intended use of these datasets. To preprocess and filter TriviaQA, CoQA, and SQuADv2 we utilized open-source code of \citep{chen_inside_2024}\footnote{\url{https://github.com/alibaba/eigenscore} (MIT license)}, which also borrows from \citep{farquhar_detecting_2024}\footnote{\url{https://github.com/lorenzkuhn/semantic_uncertainty} (MIT license)}.
In Figure \ref{fig:token_stats}, we provide histogram plots of the number of tokens for $question$ and $answer$ of each dataset computed with \texttt{meta-llama/Llama-3.1-8B-Instruct} tokenizer. 

\section{Hallucination dataset sizes}
\label{sec:ds_sizes}
\begin{figure*}[htb]
    \centering
    \includegraphics[width=\textwidth, height=0.9\textheight, keepaspectratio]{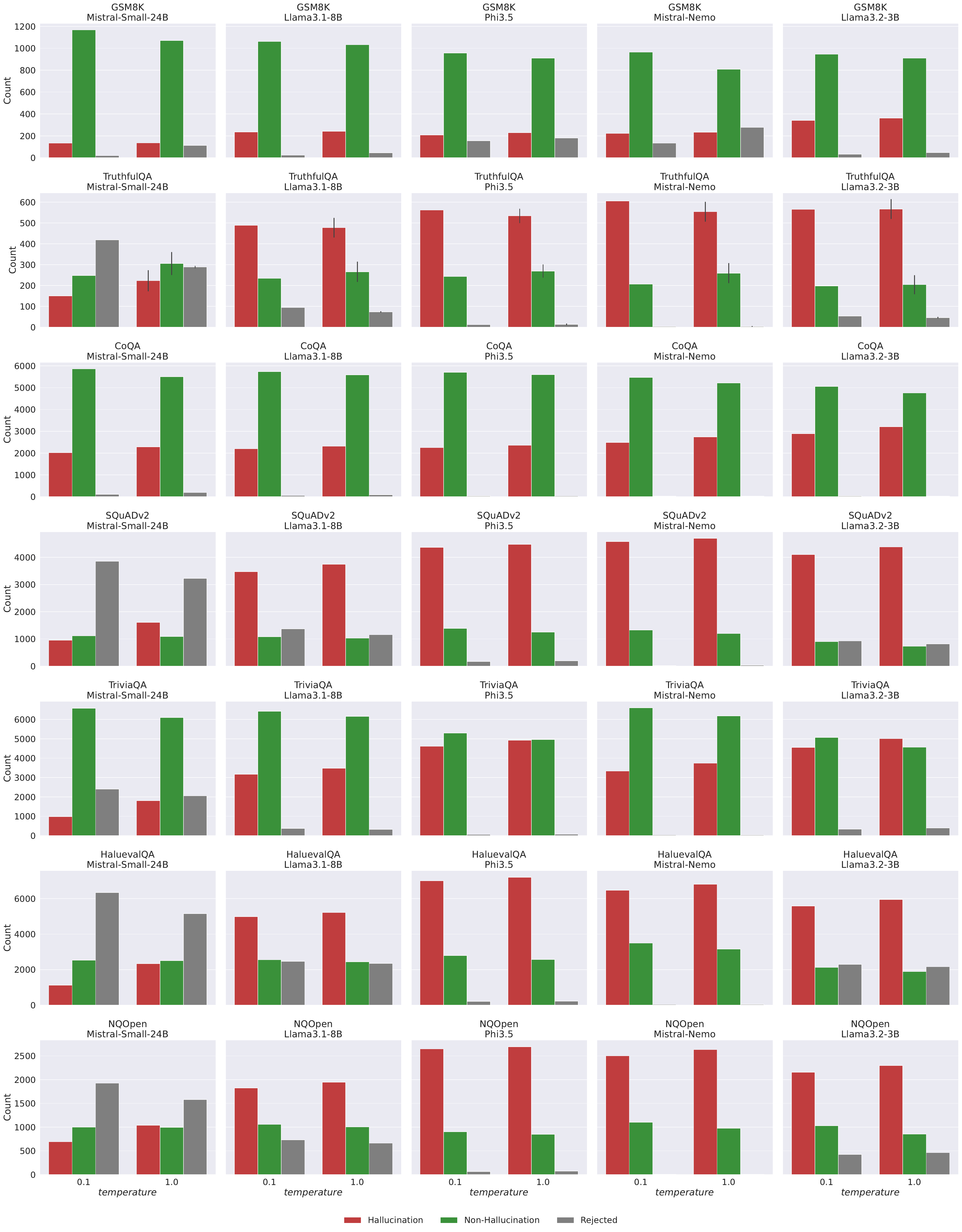}
    \caption{Number of examples per each label in generated datasets ($Hallucination$ - number of hallucinated examples, $Non{-}Hallucination$ - number of truthful examples, $Rejected$ - number of examples unable to evaluate). }
    \label{fig:ds_sizes}
\end{figure*}

Figure \ref{fig:ds_sizes} shows the number of examples per label, determined using exact match for GSM8K and the \llmjudge\ heuristic for the other datasets. It is worth noting that different generation configurations result in different splits, as LLMs might produce different answers. All examples classified as $Rejected$ were discarded from the hallucination probe training and evaluation. We observe that most datasets are imbalanced, typically underrepresenting non-hallucinated examples, with the exception of TriviaQA and GSM8K. We split each dataset into 80\% training examples and 20\% test examples. Splits were stratified according to hallucination labels.

\section{LLM-as-Judge agreement}
\label{sec:appendix_cohen_kappa}
To ensure the high quality of labels generated using the $\llmjudge$ approach, we complemented manual evaluation of random examples with a second judge LLM and measured agreement between the models. We assume that higher agreement among LLMs indicates better label quality. The reduced performance of $\lapeig$ on TriviaQA may be attributed to the lower agreement, as well as the dataset's size and class imbalance discussed earlier.

\begin{table}[ht]
\centering
\caption{Agreement between LLM judges labeling hallucinations (\gptmini, \gptnew), measured with Cohen's Kappa.}
\begin{tabular}{l c}
\hline
\textbf{Dataset} & \textbf{Cohen's Kappa} \\
\hline
CoQA        & 0.876 \\
HaluevalQA  & 0.946 \\
NQOpen      & 0.883 \\
SquadV2     & 0.854 \\
TriviaQA    & 0.939 \\
TruthfulQA  & 0.714 \\
\hline
\end{tabular}

\label{tab:kappa}
\end{table}

\section{Extended results}
\label{sec:appendix_detailed_results}

\subsection{Precision and Recall analysis}
\label{sec:appendix_extended_precision_recall}
To provide insights relevant for potential practical usage, we analyze the Precision and Recall of our method. While it has not yet been fully evaluated in production settings, this analysis illustrates the trade-offs between these metrics and informs how the method might behave in real-world applications. Metrics were computed using the default threshold of 0.5, as reported in Table \ref{tab:precision_recall}. Although trade-off patterns vary across datasets, they are consistent across all evaluated LLMs. Specifically, we observe higher recall on CoQA, GSM8K, and TriviaQA, whereas HaluEvalQA, NQ-Open, SQuADv2, and TruthfulQA exhibit higher precision. These insights can guide threshold adjustments to balance precision and recall for different production scenarios.

\begin{table*}[ht]
    \centering
    \caption{Precision and Recall values for the $\lapeig$ method, complementary to AUROC presented in Table \ref{tab:main_results}. Values are presented as Precision / Recall for each dataset and model combination.}
        \resizebox{\textwidth}{!}{%
    \begin{tabular}{lccccccc}
\toprule
\textbf{LLM} & \textbf{CoQA} & \textbf{GSM8K} & \textbf{HaluEvalQA} & \textbf{NQOpen} & \textbf{SQuADv2} & \textbf{TriviaQA} & \textbf{TruthfulQA} \\
\midrule
Llama3.1-8B        & 0.583 / 0.710 & 0.644 / 0.729 & 0.895 / 0.785 & 0.859 / 0.740 & 0.896 / 0.720 & 0.719 / 0.812 & 0.872 / 0.781 \\
Llama3.2-3B        & 0.679 / 0.728 & 0.718 / 0.699 & 0.912 / 0.788 & 0.894 / 0.662 & 0.924 / 0.720 & 0.787 / 0.729 & 0.910 / 0.746 \\
Phi3.5             & 0.560 / 0.703 & 0.600 / 0.739 & 0.899 / 0.768 & 0.910 / 0.785 & 0.906 / 0.731 & 0.787 / 0.785 & 0.829 / 0.798 \\
Mistral-Nemo       & 0.646 / 0.714 & 0.594 / 0.809 & 0.873 / 0.760 & 0.875 / 0.751 & 0.920 / 0.756 & 0.707 / 0.769 & 0.892 / 0.825 \\
Mistral-Small-24B  & 0.610 / 0.779 & 0.561 / 0.852 & 0.811 / 0.801 & 0.700 / 0.750 & 0.784 / 0.789 & 0.575 / 0.787 & 0.679 / 0.655 \\
\bottomrule
\end{tabular}
    }
    \label{tab:precision_recall}
\end{table*}

\subsection{Extended method comparison}
\label{sec:appendix_extended_method_comparison}
In Tables~\ref{tab:detailed_results} and \ref{tab:detailed_results_2}, we present the extended results corresponding to those summarized in Table~\ref{tab:main_results} in the main part of this paper. The extended results cover probes trained with both \textit{all-layers} and \textit{per-layer} variants across all models, as well as lower temperature ($temp \in \{0.1, 1.0\}$). In almost all cases, the \textit{all-layers} variant outperforms the \textit{per-layer} variant, suggesting that hallucination-related information is distributed across multiple layers. Additionally, we observe a smaller generalization gap (measured as the difference between test and training performance) for the $\lapeig$ method, indicating more robust features present in the Laplacian eigenvalues. Finally, as demonstrated in Section~\ref{fig:temperature_ablation}, increasing the temperature during answer generation improves probe performance, which is also evident in Table~\ref{tab:detailed_results}, where probes trained on answers generated with $temp{=}1.0$ consistently outperform those trained on data generated with $temp{=}0.1$.

\begin{sidewaystable*}[htb]
    \centering
    \small
    \caption{(Part I) Performance comparison of methods on an extended set of configurations. We mark results for $\attnscore$ in \textcolor{gray}{gray} as it is an unsupervised approach, not directly comparable to the others. In \textbf{bold}, we highlight the best performance on the test split of data, individually for each dataset, LLM, and temperature.}
    \label{tab:detailed_results}
    \resizebox{\textwidth}{!}{
    \begin{tabular}{lllll|rrrrrrr|rrrrrrr}
    \toprule
    LLM & Temp & Feature & \textit{all-layers} & \textit{per-layer} & \multicolumn{7}{c}{Train AUROC} & \multicolumn{7}{c}{Test AUROC} \\
    \cmidrule(lr){6-12} \cmidrule(lr){13-19}
     &  &  &  &  & CoQA & GSM8K & HaluevalQA & NQOpen & SQuADv2 & TriviaQA & TruthfulQA & CoQA & GSM8K & HaluevalQA & NQOpen & SQuADv2 & TriviaQA & TruthfulQA \\
    \midrule
    \textcolor{gray}{Llama3.1-8B} & \textcolor{gray}{0.1} & \textcolor{gray}{$\attnscore$} &  & \textcolor{gray}{\checkmark} & \textcolor{gray}{0.509} & \textcolor{gray}{0.683} & \textcolor{gray}{0.667} & \textcolor{gray}{0.607} & \textcolor{gray}{0.556} & \textcolor{gray}{0.567} & \textcolor{gray}{0.563} & \textcolor{gray}{0.541} & \textcolor{gray}{0.764} & \textcolor{gray}{0.653} & \textcolor{gray}{0.631} & \textcolor{gray}{0.575} & \textcolor{gray}{0.571} & \textcolor{gray}{0.650} \\
    \textcolor{gray}{Llama3.1-8B }& \textcolor{gray}{0.1} & \textcolor{gray}{$\attnscore$} & \textcolor{gray}{\checkmark} &  & \textcolor{gray}{0.494} & \textcolor{gray}{0.677} & \textcolor{gray}{0.614} & \textcolor{gray}{0.568} & \textcolor{gray}{0.522} & \textcolor{gray}{0.522} & \textcolor{gray}{0.489} & \textcolor{gray}{0.504} & \textcolor{gray}{0.708} & \textcolor{gray}{0.587} & \textcolor{gray}{0.558} & \textcolor{gray}{0.521} & \textcolor{gray}{0.511} & \textcolor{gray}{0.537} \\
    Llama3.1-8B & 0.1 & $\attnlogdet$ &  & \checkmark & 0.574 & 0.810 & 0.776 & 0.702 & 0.688 & 0.739 & 0.709 & 0.606 & 0.840 & 0.770 & 0.713 & 0.708 & 0.741 & 0.777 \\
    Llama3.1-8B & 0.1 & $\attnlogdet$ & \checkmark &  & 0.843 & 0.977 & 0.884 & 0.851 & 0.839 & 0.861 & 0.913 & 0.770 & 0.833 & 0.837 & 0.768 & 0.758 & 0.827 & 0.820 \\
    Llama3.1-8B & 0.1 & $\attneig$ &  & \checkmark & 0.764 & 0.879 & 0.828 & 0.713 & 0.742 & 0.793 & 0.680 & 0.729 & 0.798 & 0.799 & 0.728 & 0.749 & 0.773 & 0.790 \\
    Llama3.1-8B & 0.1 & $\attneig$ & \checkmark &  & 0.861 & 0.992 & 0.895 & 0.878 & 0.858 & 0.867 & 0.979 & 0.776 & 0.841 & 0.838 & 0.755 & 0.781 & 0.822 & 0.819 \\
    Llama3.1-8B & 0.1 & $\lapeig$ &  & \checkmark & 0.758 & 0.777 & 0.817 & 0.698 & 0.707 & 0.781 & 0.708 & 0.757 & 0.844 & 0.793 & 0.711 & 0.733 & 0.780 & 0.764 \\
    Llama3.1-8B & 0.1 & $\lapeig$ & \checkmark &  & 0.869 & 0.928 & 0.901 & 0.864 & 0.855 & 0.896 & 0.903 & \textbf{0.836} & \textbf{0.887} & \textbf{0.867} & \textbf{0.793} & \textbf{0.782} & \textbf{0.872} & \textbf{0.822} \\
    \midrule
    \textcolor{gray}{Llama3.1-8B} & \textcolor{gray}{1.0} & \textcolor{gray}{$\attnscore$} &  & \textcolor{gray}{\checkmark} & \textcolor{gray}{0.514} & \textcolor{gray}{0.705} & \textcolor{gray}{0.640} & \textcolor{gray}{0.607} & \textcolor{gray}{0.558} & \textcolor{gray}{0.578} & \textcolor{gray}{0.533} & \textcolor{gray}{0.525} & \textcolor{gray}{0.731} & \textcolor{gray}{0.642} & \textcolor{gray}{0.607} & \textcolor{gray}{0.572} & \textcolor{gray}{0.602} & \textcolor{gray}{0.629} \\
    \textcolor{gray}{Llama3.1-8B} & \textcolor{gray}{1.0} & \textcolor{gray}{$\attnscore$} & \textcolor{gray}{\checkmark} &  & \textcolor{gray}{0.507} & \textcolor{gray}{0.710} & \textcolor{gray}{0.602} & \textcolor{gray}{0.580} & \textcolor{gray}{0.534} & \textcolor{gray}{0.535} & \textcolor{gray}{0.546} & \textcolor{gray}{0.493} & \textcolor{gray}{0.720} & \textcolor{gray}{0.589} & \textcolor{gray}{0.556} & \textcolor{gray}{0.538} & \textcolor{gray}{0.532} & \textcolor{gray}{0.541} \\
    Llama3.1-8B & 1.0 & $\attnlogdet$ &  & \checkmark & 0.596 & 0.791 & 0.755 & 0.704 & 0.697 & 0.750 & 0.757 & 0.597 & 0.828 & 0.763 & 0.757 & 0.686 & 0.754 & 0.771 \\
    Llama3.1-8B & 1.0 & $\attnlogdet$ & \checkmark &  & 0.848 & 0.973 & 0.882 & 0.856 & 0.846 & 0.867 & 0.930 & 0.769 & 0.826 & 0.827 & 0.793 & 0.748 & 0.842 & 0.814 \\
    Llama3.1-8B & 1.0 & $\attneig$ &  & \checkmark & 0.762 & 0.864 & 0.820 & 0.758 & 0.754 & 0.800 & 0.796 & 0.723 & 0.812 & 0.784 & 0.732 & 0.728 & 0.796 & 0.770 \\
    Llama3.1-8B & 1.0 & $\attneig$ & \checkmark &  & 0.867 & 0.995 & 0.889 & 0.873 & 0.867 & 0.876 & 0.972 & 0.782 & 0.838 & 0.819 & 0.790 & 0.768 & 0.843 & \textbf{0.833} \\
    Llama3.1-8B & 1.0 & $\lapeig$ &  & \checkmark & 0.760 & 0.873 & 0.803 & 0.732 & 0.722 & 0.795 & 0.751 & 0.743 & 0.833 & 0.789 & 0.725 & 0.724 & 0.794 & 0.764 \\
    Llama3.1-8B & 1.0 & $\lapeig$ & \checkmark &  & 0.879 & 0.936 & 0.896 & 0.866 & 0.857 & 0.901 & 0.918 & \textbf{0.830} & \textbf{0.872} & \textbf{0.874} & \textbf{0.827} & \textbf{0.791} & \textbf{0.889} & 0.829 \\
    \midrule
    \textcolor{gray}{Llama3.2-3B} & \textcolor{gray}{0.1} & \textcolor{gray}{$\attnscore$} &  & \textcolor{gray}{\checkmark} & \textcolor{gray}{0.526} & \textcolor{gray}{0.662} & \textcolor{gray}{0.697} & \textcolor{gray}{0.592} & \textcolor{gray}{0.570} & \textcolor{gray}{0.570} & \textcolor{gray}{0.569} & \textcolor{gray}{0.547} & \textcolor{gray}{0.640} & \textcolor{gray}{0.714} & \textcolor{gray}{0.643} & \textcolor{gray}{0.582} & \textcolor{gray}{0.551} & \textcolor{gray}{0.564} \\
    \textcolor{gray}{Llama3.2-3B} & \textcolor{gray}{0.1} & \textcolor{gray}{$\attnscore$} & \textcolor{gray}{\checkmark} &  & \textcolor{gray}{0.506} & \textcolor{gray}{0.638} & \textcolor{gray}{0.635} & \textcolor{gray}{0.523} & \textcolor{gray}{0.515} & \textcolor{gray}{0.534} & \textcolor{gray}{0.473} & \textcolor{gray}{0.519} & \textcolor{gray}{0.609} & \textcolor{gray}{0.644} & \textcolor{gray}{0.573} & \textcolor{gray}{0.561} & \textcolor{gray}{0.510} & \textcolor{gray}{0.489} \\
    Llama3.2-3B & 0.1 & $\attnlogdet$ &  & \checkmark & 0.573 & 0.774 & 0.762 & 0.692 & 0.682 & 0.719 & 0.725 & 0.579 & 0.794 & 0.774 & 0.735 & 0.698 & 0.711 & 0.674 \\
    Llama3.2-3B & 0.1 & $\attnlogdet$ & \checkmark &  & 0.782 & 0.946 & 0.868 & 0.845 & 0.827 & 0.824 & 0.918 & 0.695 & 0.841 & 0.843 & 0.763 & \textbf{0.749} & 0.796 & 0.678 \\
    Llama3.2-3B & 0.1 & $\attneig$ &  & \checkmark & 0.675 & 0.784 & 0.782 & 0.750 & 0.725 & 0.755 & 0.727 & 0.626 & 0.761 & 0.792 & 0.734 & 0.695 & 0.724 & 0.720 \\
    Llama3.2-3B & 0.1 & $\attneig$ & \checkmark &  & 0.814 & 0.977 & 0.873 & 0.872 & 0.852 & 0.842 & 0.963 & 0.723 & 0.808 & 0.844 & 0.772 & 0.744 & 0.788 & 0.688 \\
    Llama3.2-3B & 0.1 & $\lapeig$ &  & \checkmark & 0.681 & 0.763 & 0.774 & 0.733 & 0.708 & 0.733 & 0.722 & 0.676 & 0.835 & 0.781 & 0.736 & 0.697 & 0.732 & 0.690 \\
    Llama3.2-3B & 0.1 & $\lapeig$ & \checkmark &  & 0.831 & 0.889 & 0.875 & 0.837 & 0.832 & 0.852 & 0.895 & \textbf{0.801} & \textbf{0.852} & \textbf{0.857} & \textbf{0.779} & 0.736 & \textbf{0.826} & \textbf{0.743} \\
    \midrule
    \textcolor{gray}{Llama3.2-3B} & \textcolor{gray}{1.0} & \textcolor{gray}{$\attnscore$} &  & \textcolor{gray}{\checkmark} & \textcolor{gray}{0.532} & \textcolor{gray}{0.674} & \textcolor{gray}{0.668} & \textcolor{gray}{0.588} & \textcolor{gray}{0.578} & \textcolor{gray}{0.553} & \textcolor{gray}{0.555} & \textcolor{gray}{0.557} & \textcolor{gray}{0.753} & \textcolor{gray}{0.637} & \textcolor{gray}{0.592} & \textcolor{gray}{0.593} & \textcolor{gray}{0.558} & \textcolor{gray}{0.675} \\
    \textcolor{gray}{Llama3.2-3B} & \textcolor{gray}{1.0} & \textcolor{gray}{$\attnscore$} & \textcolor{gray}{\checkmark} &  & \textcolor{gray}{0.512} & \textcolor{gray}{0.648} & \textcolor{gray}{0.606} & \textcolor{gray}{0.554} & \textcolor{gray}{0.529} & \textcolor{gray}{0.517} & \textcolor{gray}{0.484} & \textcolor{gray}{0.509} & \textcolor{gray}{0.717} & \textcolor{gray}{0.588} & \textcolor{gray}{0.546} & \textcolor{gray}{0.530} & \textcolor{gray}{0.515} & \textcolor{gray}{0.581} \\
    Llama3.2-3B & 1.0 & $\attnlogdet$ &  & \checkmark & 0.578 & 0.807 & 0.738 & 0.677 & 0.720 & 0.716 & 0.739 & 0.597 & 0.816 & 0.724 & 0.678 & 0.707 & 0.711 & 0.742 \\
    Llama3.2-3B & 1.0 & $\attnlogdet$ & \checkmark &  & 0.784 & 0.951 & 0.869 & 0.816 & 0.839 & 0.831 & 0.924 & 0.700 & 0.851 & 0.801 & 0.690 & 0.734 & 0.789 & \textbf{0.795} \\
    Llama3.2-3B & 1.0 & $\attneig$ &  & \checkmark & 0.642 & 0.807 & 0.777 & 0.716 & 0.747 & 0.763 & 0.735 & 0.641 & 0.817 & 0.756 & 0.696 & 0.703 & 0.746 & 0.748 \\
    Llama3.2-3B & 1.0 & $\attneig$ & \checkmark &  & 0.819 & 0.973 & 0.878 & 0.847 & 0.876 & 0.847 & 0.978 & 0.724 & 0.768 & 0.819 & \textbf{0.694} & 0.749 & 0.804 & 0.723 \\
    Llama3.2-3B & 1.0 & $\lapeig$ &  & \checkmark & 0.695 & 0.781 & 0.764 & 0.683 & 0.719 & 0.727 & 0.682 & 0.715 & 0.815 & 0.754 & 0.671 & 0.711 & 0.738 & 0.767 \\
    Llama3.2-3B & 1.0 & $\lapeig$ & \checkmark &  & 0.842 & 0.894 & 0.885 & 0.803 & 0.850 & 0.863 & 0.911 & \textbf{0.812} & \textbf{0.870 }& \textbf{0.828} & 0.693 & \textbf{0.757} & \textbf{0.832} & 0.787 \\
    \midrule
    \textcolor{gray}{Phi3.5} & \textcolor{gray}{0.1} & \textcolor{gray}{$\attnscore$} &  & \textcolor{gray}{\checkmark} & \textcolor{gray}{0.517} & \textcolor{gray}{0.723} & \textcolor{gray}{0.559} & \textcolor{gray}{0.565} & \textcolor{gray}{0.606} & \textcolor{gray}{0.625} & \textcolor{gray}{0.601} & \textcolor{gray}{0.528} & \textcolor{gray}{0.682} & \textcolor{gray}{0.551} & \textcolor{gray}{0.637} & \textcolor{gray}{0.621} & \textcolor{gray}{0.628} & \textcolor{gray}{0.637} \\
    \textcolor{gray}{Phi3.5} & \textcolor{gray}{0.1} & \textcolor{gray}{$\attnscore$} & \textcolor{gray}{\checkmark} &  & \textcolor{gray}{0.499} & \textcolor{gray}{0.632} & \textcolor{gray}{0.538} & \textcolor{gray}{0.532} & \textcolor{gray}{0.473} & \textcolor{gray}{0.539} & \textcolor{gray}{0.522} & \textcolor{gray}{0.505} & \textcolor{gray}{0.605} & \textcolor{gray}{0.511} & \textcolor{gray}{0.578} & \textcolor{gray}{0.458} & \textcolor{gray}{0.534} & \textcolor{gray}{0.554} \\
    Phi3.5 & 0.1 & $\attnlogdet$ &  & \checkmark & 0.583 & 0.805 & 0.732 & 0.741 & 0.711 & 0.757 & 0.720 & 0.585 & 0.749 & 0.726 & 0.785 & 0.726 & 0.772 & 0.765 \\
    Phi3.5 & 0.1 & $\attnlogdet$ & \checkmark &  & 0.845 & 0.995 & 0.863 & 0.905 & 0.852 & 0.875 & 0.981 & 0.723 & 0.752 & 0.802 & 0.802 & 0.759 & 0.842 & 0.716 \\
    Phi3.5 & 0.1 & $\attneig$ &  & \checkmark & 0.760 & 0.882 & 0.781 & 0.793 & 0.745 & 0.802 & 0.854 & 0.678 & 0.764 & 0.764 & 0.790 & 0.747 & 0.791 & \textbf{0.774} \\
    Phi3.5 & 0.1 & $\attneig$ & \checkmark &  & 0.862 & 1.000 & 0.867 & 0.904 & 0.861 & 0.881 & 0.999 & 0.728 & 0.732 & 0.802 & 0.787 & 0.740 & 0.838 & 0.761 \\
    Phi3.5 & 0.1 & $\lapeig$ &  & \checkmark & 0.734 & 0.713 & 0.758 & 0.737 & 0.704 & 0.775 & 0.759 & 0.716 & 0.753 & 0.757 & 0.761 & 0.732 & 0.768 & 0.741 \\
    Phi3.5 & 0.1 & $\lapeig$ & \checkmark &  & 0.856 & 0.946 & 0.860 & 0.897 & 0.841 & 0.884 & 0.965 & \textbf{0.810} & \textbf{0.785} & \textbf{0.819} & \textbf{0.815} & \textbf{0.791} & \textbf{0.858} & 0.717 \\
    \midrule
    \textcolor{gray}{Phi3.5} & \textcolor{gray}{1.0} & \textcolor{gray}{$\attnscore$} &  & \textcolor{gray}{\checkmark} & \textcolor{gray}{0.499} & \textcolor{gray}{0.699} & \textcolor{gray}{0.567} & \textcolor{gray}{0.615} & \textcolor{gray}{0.626} & \textcolor{gray}{0.637} & \textcolor{gray}{0.618} & \textcolor{gray}{0.533} & \textcolor{gray}{0.722} & \textcolor{gray}{0.581} & \textcolor{gray}{0.630} & \textcolor{gray}{0.645} & \textcolor{gray}{0.642} & \textcolor{gray}{0.626} \\
    \textcolor{gray}{Phi3.5} & \textcolor{gray}{1.0} & \textcolor{gray}{$\attnscore$} & \textcolor{gray}{\checkmark} &  & \textcolor{gray}{0.489} & \textcolor{gray}{0.640} & \textcolor{gray}{0.540} & \textcolor{gray}{0.566} & \textcolor{gray}{0.469} & \textcolor{gray}{0.553} & \textcolor{gray}{0.541} & \textcolor{gray}{0.520} & \textcolor{gray}{0.666} & \textcolor{gray}{0.541} & \textcolor{gray}{0.594} & \textcolor{gray}{0.504} & \textcolor{gray}{0.540} & \textcolor{gray}{0.554} \\
    Phi3.5 & 1.0 & $\attnlogdet$ &  & \checkmark & 0.587 & 0.831 & 0.733 & 0.773 & 0.722 & 0.766 & 0.753 & 0.557 & 0.842 & 0.762 & 0.784 & 0.736 & 0.772 & 0.763 \\
    Phi3.5 & 1.0 & $\attnlogdet$ & \checkmark &  & 0.842 & 0.993 & 0.868 & 0.921 & 0.859 & 0.879 & 0.971 & 0.745 & 0.842 & 0.818 & 0.815 & 0.769 & 0.848 & 0.755 \\
    Phi3.5 & 1.0 & $\attneig$ &  & \checkmark & 0.755 & 0.852 & 0.794 & 0.820 & 0.790 & 0.809 & 0.864 & 0.710 & 0.809 & 0.795 & 0.787 & 0.752 & 0.799 & 0.747 \\
    Phi3.5 & 1.0 & $\attneig$ & \checkmark &  & 0.858 & 1.000 & 0.871 & 0.924 & 0.876 & 0.887 & 0.998 & 0.771 & 0.794 & 0.829 & 0.798 & 0.782 & 0.850 & \textbf{0.802} \\
    Phi3.5 & 1.0 & $\lapeig$ &  & \checkmark & 0.733 & 0.771 & 0.755 & 0.755 & 0.718 & 0.779 & 0.713 & 0.723 & 0.816 & 0.769 & 0.755 & 0.732 & 0.792 & 0.732 \\
    Phi3.5 & 1.0 & $\lapeig$ & \checkmark &  & 0.856 & 0.937 & 0.863 & 0.911 & 0.849 & 0.889 & 0.961 & \textbf{0.821} & \textbf{0.885} & \textbf{0.836} & \textbf{0.826} & \textbf{0.795} & \textbf{0.872} & 0.777 \\
    \bottomrule
    
    \end{tabular}
    }
\end{sidewaystable*}

\begin{sidewaystable*}[htb]
    \centering
    \small
    \caption{(Part II) Performance comparison of methods on an extended set of configurations. We mark results for $\attnscore$ in \textcolor{gray}{gray} as it is an unsupervised approach, not directly comparable to the others. In \textbf{bold}, we highlight the best performance on the test split of data, individually for each dataset, LLM, and temperature.}
    \label{tab:detailed_results_2}
    \resizebox{\textwidth}{!}{
    \begin{tabular}{lllll|rrrrrrr|rrrrrrr}
     \toprule
     LLM & Temp & Feature & \textit{all-layers} & \textit{per-layer} & \multicolumn{7}{c}{Train AUROC} & \multicolumn{7}{c}{Test AUROC} \\
     \cmidrule(lr){6-12} \cmidrule(lr){13-19}
      &  &  &  &  & CoQA & GSM8K & HaluevalQA & NQOpen & SQuADv2 & TriviaQA & TruthfulQA & CoQA & GSM8K& HaluevalQA & NQOpen & SQuADv2 & TriviaQA & TruthfulQA \\
    \midrule
    \textcolor{gray}{Mistral-Nemo} & \textcolor{gray}{0.1} & \textcolor{gray}{$\attnscore$} & \textcolor{gray}{} & \textcolor{gray}{\checkmark} & \textcolor{gray}{0.504} & \textcolor{gray}{0.727} & \textcolor{gray}{0.574} & \textcolor{gray}{0.591} & \textcolor{gray}{0.509} & \textcolor{gray}{0.550} & \textcolor{gray}{0.546} & \textcolor{gray}{0.515} & \textcolor{gray}{0.697} & \textcolor{gray}{0.559} & \textcolor{gray}{0.587} & \textcolor{gray}{0.527} & \textcolor{gray}{0.545} & \textcolor{gray}{0.681} \\
    \textcolor{gray}{Mistral-Nemo} & \textcolor{gray}{0.1} & \textcolor{gray}{$\attnscore$} & \textcolor{gray}{\checkmark} & \textcolor{gray}{} & \textcolor{gray}{0.508} & \textcolor{gray}{0.707} & \textcolor{gray}{0.536} & \textcolor{gray}{0.537} & \textcolor{gray}{0.507} & \textcolor{gray}{0.520} & \textcolor{gray}{0.535} & \textcolor{gray}{0.484} & \textcolor{gray}{0.667} & \textcolor{gray}{0.523} & \textcolor{gray}{0.533} & \textcolor{gray}{0.495} & \textcolor{gray}{0.505} & \textcolor{gray}{0.631} \\
    Mistral-Nemo & 0.1 & $\attnlogdet$ &  &  \checkmark & 0.584 & 0.801 & 0.716 & 0.702 & 0.675 & 0.689 & 0.744 & 0.583 & 0.807 & 0.723 & 0.688 & 0.668 & 0.722 & 0.731 \\
    Mistral-Nemo & 0.1 & $\attnlogdet$ & \checkmark &  & 0.828 & 0.993 & 0.842 & 0.861 & 0.858 & 0.854 & 0.963 & 0.734 & \textbf{0.820} & 0.786 & 0.752 & 0.709 & 0.822 & \textbf{0.776} \\
    Mistral-Nemo & 0.1 & $\attneig$ &  &  \checkmark & 0.708 & 0.865 & 0.751 & 0.749 & 0.749 & 0.747 & 0.797 & 0.672 & 0.795 & 0.740 & 0.701 & 0.704 & 0.738 & 0.717 \\
    Mistral-Nemo & 0.1 & $\attneig$ & \checkmark &  & 0.845 & 1.000 & 0.842 & 0.878 & 0.864 & 0.859 & 0.996 & 0.768 & 0.771 & 0.789 & 0.743 & 0.716 & 0.809 & 0.752 \\
    Mistral-Nemo & 0.1 & $\lapeig$ &  &  \checkmark & 0.763 & 0.777 & 0.772 & 0.732 & 0.723 & 0.781 & 0.725 & 0.759 & 0.751 & 0.760 & 0.697 & 0.696 & 0.769 & 0.710 \\
    Mistral-Nemo & 0.1 & $\lapeig$ & \checkmark &  & 0.868 & 0.969 & 0.862 & 0.875 & 0.869 & 0.886 & 0.977 & \textbf{0.823} & 0.805 & \textbf{0.821} & \textbf{0.755} & \textbf{0.767} & \textbf{0.858} & 0.737 \\
    \midrule
    \textcolor{gray}{Mistral-Nemo} & \textcolor{gray}{1.0} & \textcolor{gray}{$\attnscore$} & \textcolor{gray}{} & \textcolor{gray}{\checkmark} & \textcolor{gray}{0.502} & \textcolor{gray}{0.656} & \textcolor{gray}{0.586} & \textcolor{gray}{0.606} & \textcolor{gray}{0.546} & \textcolor{gray}{0.553} & \textcolor{gray}{0.570} & \textcolor{gray}{0.525} & \textcolor{gray}{0.670} & \textcolor{gray}{0.587} & \textcolor{gray}{0.588} & \textcolor{gray}{0.564} & \textcolor{gray}{0.570} & \textcolor{gray}{0.632} \\
    \textcolor{gray}{Mistral-Nemo} & \textcolor{gray}{1.0} & \textcolor{gray}{$\attnscore$} & \textcolor{gray}{\checkmark} & \textcolor{gray}{} & \textcolor{gray}{0.493} & \textcolor{gray}{0.675} & \textcolor{gray}{0.541} & \textcolor{gray}{0.552} & \textcolor{gray}{0.503} & \textcolor{gray}{0.521} & \textcolor{gray}{0.531} & \textcolor{gray}{0.493} & \textcolor{gray}{0.630} & \textcolor{gray}{0.531} & \textcolor{gray}{0.529} & \textcolor{gray}{0.510} & \textcolor{gray}{0.532} & \textcolor{gray}{0.494} \\
    Mistral-Nemo & 1.0 & $\attnlogdet$ &  &  \checkmark & 0.591 & 0.790 & 0.723 & 0.716 & 0.717 & 0.717 & 0.741 & 0.581 & 0.782 & 0.730 & 0.703 & 0.711 & 0.707 & 0.801 \\
    Mistral-Nemo & 1.0 & $\attnlogdet$ & \checkmark &  & 0.829 & 0.994 & 0.851 & 0.870 & 0.860 & 0.857 & 0.963 & 0.728 & 0.856 & 0.798 & 0.769 & 0.772 & 0.812 & \textbf{0.852} \\
    Mistral-Nemo & 1.0 & $\attneig$ &  &  \checkmark & 0.704 & 0.845 & 0.762 & 0.742 & 0.757 & 0.752 & 0.806 & 0.670 & 0.781 & 0.749 & 0.742 & 0.719 & 0.737 & 0.804 \\
    Mistral-Nemo & 1.0 & $\attneig$ & \checkmark &  & 0.844 & 1.000 & 0.851 & 0.893 & 0.864 & 0.862 & 0.996 & 0.778 & 0.842 & 0.781 & 0.761 & 0.758 & 0.821 & 0.802 \\
    Mistral-Nemo & 1.0 & $\lapeig$ &  &  \checkmark & 0.765 & 0.820 & 0.790 & 0.749 & 0.740 & 0.804 & 0.779 & 0.738 & 0.808 & 0.763 & 0.708 & 0.723 & 0.785 & 0.818 \\
    Mistral-Nemo & 1.0 & $\lapeig$ & \checkmark &  & 0.876 & 0.965 & 0.877 & 0.884 & 0.881 & 0.901 & 0.978 & \textbf{0.835} & \textbf{0.890} & \textbf{0.833} & \textbf{0.795} & \textbf{0.812} & \textbf{0.865} & 0.828 \\
    \midrule
    \textcolor{gray}{Mistral-Small-24B} & \textcolor{gray}{0.1} & \textcolor{gray}{$\attnscore$} & \textcolor{gray}{} & \textcolor{gray}{\checkmark} & \textcolor{gray}{0.520} & \textcolor{gray}{0.759} & \textcolor{gray}{0.538} & \textcolor{gray}{0.517} & \textcolor{gray}{0.577} & \textcolor{gray}{0.535} & \textcolor{gray}{0.571} & \textcolor{gray}{0.525} & \textcolor{gray}{0.685} & \textcolor{gray}{0.552} & \textcolor{gray}{0.592} & \textcolor{gray}{0.625} & \textcolor{gray}{0.533} & \textcolor{gray}{0.724} \\
    \textcolor{gray}{Mistral-Small-24B} & \textcolor{gray}{0.1} & \textcolor{gray}{$\attnscore$} & \textcolor{gray}{\checkmark} & \textcolor{gray}{} & \textcolor{gray}{0.520} & \textcolor{gray}{0.668} & \textcolor{gray}{0.472} & \textcolor{gray}{0.449} & \textcolor{gray}{0.510} & \textcolor{gray}{0.449} & \textcolor{gray}{0.491} & \textcolor{gray}{0.493} & \textcolor{gray}{0.578} & \textcolor{gray}{0.493} & \textcolor{gray}{0.467} & \textcolor{gray}{0.556} & \textcolor{gray}{0.461} & \textcolor{gray}{0.645} \\
    Mistral-Small-24B & 0.1 & $\attnlogdet$ &  &  \checkmark & 0.585 & 0.834 & 0.674 & 0.659 & 0.724 & 0.685 & 0.698 & 0.586 & 0.809 & 0.684 & 0.695 & 0.752 & 0.682 & 0.721 \\
    Mistral-Small-24B & 0.1 & $\attnlogdet$ & \checkmark &  & 0.851 & 0.990 & 0.817 & 0.799 & 0.820 & 0.861 & 0.898 & 0.762 & \textbf{0.896} & 0.760 & 0.725 & 0.763 & 0.778 & 0.767 \\
    Mistral-Small-24B & 0.1 & $\attneig$ &  &  \checkmark & 0.734 & 0.863 & 0.722 & 0.667 & 0.745 & 0.757 & 0.732 & 0.720 & 0.837 & 0.707 & 0.697 & 0.773 & 0.758 & 0.765 \\
    Mistral-Small-24B & 0.1 & $\attneig$ & \checkmark &  & 0.872 & 0.999 & 0.873 & 0.923 & 0.903 & 0.899 & 0.993 & 0.793 & \textbf{0.896} & 0.771 & \textbf{0.731} & 0.803 & 0.809 & 0.796 \\
    Mistral-Small-24B & 0.1 & $\lapeig$ &  &  \checkmark & 0.802 & 0.781 & 0.720 & 0.646 & 0.714 & 0.742 & 0.694 & 0.800 & 0.850 & 0.719 & 0.674 & 0.784 & 0.757 & \textbf{0.827} \\
    Mistral-Small-24B & 0.1 & $\lapeig$ & \checkmark &  & 0.887 & 0.985 & 0.870 & 0.901 & 0.887 & 0.905 & 0.979 & \textbf{0.852} & 0.881 & \textbf{0.808} & 0.722 & \textbf{0.821} & \textbf{0.831} & 0.757 \\
    \midrule
    \textcolor{gray}{Mistral-Small-24B} & \textcolor{gray}{1.0} & \textcolor{gray}{$\attnscore$} & \textcolor{gray}{} & \textcolor{gray}{\checkmark} & \textcolor{gray}{0.511} & \textcolor{gray}{0.706} & \textcolor{gray}{0.555} & \textcolor{gray}{0.582} & \textcolor{gray}{0.561} & \textcolor{gray}{0.562} & \textcolor{gray}{0.542} & \textcolor{gray}{0.535} & \textcolor{gray}{0.713} & \textcolor{gray}{0.566} & \textcolor{gray}{0.576} & \textcolor{gray}{0.567} & \textcolor{gray}{0.574} & \textcolor{gray}{0.606} \\
    \textcolor{gray}{Mistral-Small-24B} & \textcolor{gray}{1.0} & \textcolor{gray}{$\attnscore$} & \textcolor{gray}{\checkmark} & \textcolor{gray}{} & \textcolor{gray}{0.497} & \textcolor{gray}{0.595} & \textcolor{gray}{0.503} & \textcolor{gray}{0.463} & \textcolor{gray}{0.519} & \textcolor{gray}{0.451} & \textcolor{gray}{0.493} & \textcolor{gray}{0.516} & \textcolor{gray}{0.576} & \textcolor{gray}{0.504} & \textcolor{gray}{0.462} & \textcolor{gray}{0.455} & \textcolor{gray}{0.463} & \textcolor{gray}{0.451} \\
    Mistral-Small-24B & 1.0 & $\attnlogdet$ &  &  \checkmark & 0.591 & 0.824 & 0.727 & 0.710 & 0.732 & 0.720 & 0.677 & 0.600 & 0.869 & 0.771 & 0.714 & 0.726 & 0.734 & 0.687 \\
    Mistral-Small-24B & 1.0 & $\attnlogdet$ & \checkmark &  & 0.850 & 0.989 & 0.847 & 0.827 & 0.856 & 0.853 & 0.877 & 0.766 & 0.853 & 0.842 & 0.747 & 0.753 & 0.833 & 0.735 \\
    Mistral-Small-24B & 1.0 & $\attneig$ &  &  \checkmark & 0.757 & 0.920 & 0.743 & 0.728 & 0.764 & 0.779 & 0.741 & 0.723 & 0.868 & 0.780 & 0.733 & 0.734 & 0.780 & 0.718 \\
    Mistral-Small-24B & 1.0 & $\attneig$ & \checkmark &  & 0.877 & 1.000 & 0.878 & 0.923 & 0.911 & 0.895 & 0.997 & 0.805 & 0.846 & 0.848 & 0.751 & 0.760 & 0.844 & \textbf{0.765} \\
    Mistral-Small-24B & 1.0 & $\lapeig$ &  &  \checkmark & 0.814 & 0.860 & 0.762 & 0.733 & 0.790 & 0.766 & 0.703 & 0.805 & 0.897 & 0.790 & 0.712 & 0.781 & 0.779 & 0.725 \\
    Mistral-Small-24B & 1.0 & $\lapeig$ & \checkmark &  & 0.895 & 0.980 & 0.890 & 0.898 & 0.910 & 0.907 & 0.965 & \textbf{0.861} & \textbf{0.925} & \textbf{0.882} & \textbf{0.791} & \textbf{0.820} & \textbf{0.876} & 0.748 \\ 
    \bottomrule
    
    \end{tabular}
    }
\end{sidewaystable*}

\subsection{Best found hyperparameters}
We present the hyperparameter values corresponding to the results in Table~\ref{tab:main_results} and Table~\ref{tab:detailed_results}. Table~\ref{tab:best_top_k} shows the optimal hyperparameter $k$ for selecting the top-$k$ eigenvalues from either the attention maps in $\attneig$ or the Laplacian matrix in $\lapeig$. While fewer eigenvalues were sufficient for optimal performance in some cases, the best results were generally achieved with the highest tested value, $k{=}100$.

Table~\ref{tab:best_layer_idx} reports the layer indices that yielded the highest performance for the \textit{per-layer} models. Performance typically peaked in layers above the 10th, especially for \llamabig, where attention maps from the final layers more often led to better hallucination detection. Interestingly, the first layer's attention maps also produced strong performance in a few cases. Overall, no clear pattern emerges regarding the optimal layer, and as noted in prior work, selecting the best layer in the \textit{per-layer} setup often requires a search.

\begin{table*}[htb]
    \centering
    \caption{Values of $k$ hyperparameter, denoting how many highest eigenvalues are taken from the Laplacian matrix, corresponding to the best results in Table~\ref{tab:main_results} and Table~\ref{tab:detailed_results}.}
    \resizebox{\textwidth}{!}{
        \begin{tabular}{lrlllrrrrrrr}
\toprule
LLM & Temp & Feature & \textit{all-layers} & \textit{per-layer} & \multicolumn{7}{c}{top-$k$ eigenvalues} \\
\cmidrule(lr){6-12}
& & & & & CoQA & GSM8K & HaluevalQA & NQOpen & SQuADv2 & TriviaQA & TruthfulQA \\
\midrule
Llama3.1-8B & 0.1 & $\attneig$ &  & \checkmark & 50 & 100 & 100 & 25 & 100 & 100 & 10 \\
Llama3.1-8B & 0.1 & $\attneig$ & \checkmark &  & 100 & 100 & 100 & 100 & 100 & 50 & 100 \\
Llama3.1-8B & 0.1 & $\lapeig$ &  & \checkmark & 50 & 50 & 100 & 10 & 100 & 100 & 100 \\
Llama3.1-8B & 0.1 & $\lapeig$ & \checkmark &  & 10 & 100 & 100 & 100 & 100 & 100 & 100 \\
\midrule
Llama3.1-8B & 1.0 & $\attneig$ &  & \checkmark & 100 & 100 & 100 & 100 & 100 & 100 & 100 \\
Llama3.1-8B & 1.0 & $\attneig$ & \checkmark &  & 100 & 100 & 100 & 100 & 100 & 100 & 100 \\
Llama3.1-8B & 1.0 & $\lapeig$ &  & \checkmark & 100 & 50 & 100 & 100 & 100 & 100 & 100 \\
Llama3.1-8B & 1.0 & $\lapeig$ & \checkmark &  & 100 & 100 & 25 & 100 & 100 & 100 & 100 \\
\midrule
Llama3.2-3B & 0.1 & $\attneig$ &  & \checkmark & 100 & 100 & 100 & 100 & 100 & 100 & 10 \\
Llama3.2-3B & 0.1 & $\attneig$ & \checkmark &  & 100 & 100 & 25 & 100 & 100 & 100 & 100 \\
Llama3.2-3B & 0.1 & $\lapeig$ &  & \checkmark & 100 & 25 & 100 & 100 & 100 & 50 & 5 \\
Llama3.2-3B & 0.1 & $\lapeig$ & \checkmark &  & 25 & 100 & 100 & 100 & 100 & 100 & 100 \\
\midrule
Llama3.2-3B & 1.0 & $\attneig$ &  & \checkmark & 100 & 100 & 100 & 100 & 100 & 100 & 50 \\
Llama3.2-3B & 1.0 & $\attneig$ & \checkmark &  & 100 & 50 & 100 & 100 & 100 & 100 & 100 \\
Llama3.2-3B & 1.0 & $\lapeig$ &  & \checkmark & 100 & 50 & 100 & 10 & 100 & 100 & 25 \\
Llama3.2-3B & 1.0 & $\lapeig$ & \checkmark &  & 25 & 100 & 100 & 100 & 100 & 100 & 100 \\
\midrule
Phi3.5 & 0.1 & $\attneig$ &  & \checkmark & 100 & 100 & 100 & 100 & 100 & 100 & 100 \\
Phi3.5 & 0.1 & $\attneig$ & \checkmark &  & 100 & 25 & 10 & 10 & 25 & 100 & 50 \\
Phi3.5 & 0.1 & $\lapeig$ &  & \checkmark & 100 & 10 & 100 & 100 & 100 & 100 & 100 \\
Phi3.5 & 0.1 & $\lapeig$ & \checkmark &  & 10 & 100 & 50 & 100 & 100 & 100 & 100 \\
\midrule
Phi3.5 & 1.0 & $\attneig$ &  & \checkmark & 100 & 100 & 100 & 100 & 100 & 100 & 100 \\
Phi3.5 & 1.0 & $\attneig$ & \checkmark &  & 100 & 100 & 100 & 10 & 100 & 100 & 50 \\
Phi3.5 & 1.0 & $\lapeig$ &  & \checkmark & 100 & 25 & 100 & 100 & 100 & 100 & 50 \\
Phi3.5 & 1.0 & $\lapeig$ & \checkmark &  & 10 & 25 & 100 & 100 & 100 & 100 & 100 \\
\midrule
Mistral-Nemo & 0.1 & $\attneig$ &  & \checkmark & 100 & 50 & 100 & 100 & 100 & 100 & 100 \\
Mistral-Nemo & 0.1 & $\attneig$ & \checkmark &  & 100 & 50 & 100 & 100 & 100 & 100 & 100 \\
Mistral-Nemo & 0.1 & $\lapeig$ &  & \checkmark & 100 & 25 & 100 & 100 & 100 & 100 & 10 \\
Mistral-Nemo & 0.1 & $\lapeig$ & \checkmark &  & 10 & 100 & 25 & 100 & 50 & 100 & 100 \\
\midrule
Mistral-Nemo & 1.0 & $\attneig$ &  & \checkmark & 100 & 100 & 100 & 100 & 100 & 100 & 100 \\
Mistral-Nemo & 1.0 & $\attneig$ & \checkmark &  & 100 & 100 & 100 & 100 & 100 & 50 & 100 \\
Mistral-Nemo & 1.0 & $\lapeig$ &  & \checkmark & 100 & 100 & 100 & 50 & 100 & 100 & 100 \\
Mistral-Nemo & 1.0 & $\lapeig$ & \checkmark &  & 10 & 100 & 50 & 100 & 100 & 100 & 100 \\
\midrule
Mistral-Small-24B & 0.1 & $\attneig$ &  & \checkmark & 100 & 100 & 100 & 10 & 100 & 50 & 25 \\
Mistral-Small-24B & 0.1 & $\attneig$ & \checkmark &  & 100 & 100 & 100 & 100 & 100 & 100 & 25 \\
Mistral-Small-24B & 0.1 & $\lapeig$ &  & \checkmark & 100 & 50 & 100 & 50 & 100 & 100 & 10 \\
Mistral-Small-24B & 0.1 & $\lapeig$ & \checkmark &  & 25 & 100 & 100 & 100 & 100 & 10 & 100 \\
\midrule
Mistral-Small-24B & 1.0 & $\attneig$ &  & \checkmark & 100 & 100 & 100 & 100 & 100 & 100 & 100 \\
Mistral-Small-24B & 1.0 & $\attneig$ & \checkmark &  & 100 & 100 & 100 & 100 & 100 & 100 & 100 \\
Mistral-Small-24B & 1.0 & $\lapeig$ &  & \checkmark & 100 & 100 & 100 & 100 & 50 & 100 & 50 \\
Mistral-Small-24B & 1.0 & $\lapeig$ & \checkmark &  & 10 & 100 & 50 & 10 & 10 & 100 & 50 \\
\bottomrule
\end{tabular}

    }
    \label{tab:best_top_k}
\end{table*}

\begin{table*}[htb]
    \centering
    \caption{Values of a layer index (numbered from 0) corresponding to the best results for \textit{per-layer} models in Table~\ref{tab:detailed_results}.}
    \resizebox{\textwidth}{!}{
    \begin{tabular}{lrlrrrrrrr}
\toprule
LLM & $temp$ & Feature & \multicolumn{7}{c}{Layer index} \\
\cmidrule(lr){4-10}
& & & CoQA & GSM8K & HaluevalQA & NQOpen & SQuADv2 & TriviaQA & TruthfulQA \\
    \midrule
    Llama3.1-8B & 0.1 & $\attnscore$ & 13 & 28 & 10 & 0 & 0 & 0 & 28 \\
    Llama3.1-8B & 0.1 & $\attnlogdet$ & 7 & 31 & 13 & 16 & 11 & 29 & 21 \\
    Llama3.1-8B & 0.1 & $\attneig$ & 22 & 31 & 31 & 26 & 31 & 31 & 7 \\
    Llama3.1-8B & 0.1 & $\lapeig$ & 15 & 25 & 14 & 20 & 29 & 31 & 20 \\
    \midrule
    Llama3.1-8B & 1.0 & $\attnscore$ & 29 & 3 & 10 & 0 & 0 & 0 & 23 \\
    Llama3.1-8B & 1.0 & $\attnlogdet$ & 17 & 16 & 11 & 13 & 29 & 29 & 30 \\
    Llama3.1-8B & 1.0 & $\attneig$ & 22 & 28 & 31 & 31 & 31 & 31 & 31 \\
    Llama3.1-8B & 1.0 & $\lapeig$ & 15 & 11 & 14 & 31 & 29 & 29 & 29 \\
    \midrule
    Llama3.2-3B & 0.1 & $\attnscore$ & 15 & 17 & 12 & 12 & 12 & 21 & 14 \\
    Llama3.2-3B & 0.1 & $\attnlogdet$ & 12 & 18 & 13 & 24 & 10 & 25 & 14 \\
    Llama3.2-3B & 0.1 & $\attneig$ & 27 & 14 & 14 & 14 & 25 & 27 & 17 \\
    Llama3.2-3B & 0.1 & $\lapeig$ & 11 & 24 & 8 & 12 & 25 & 12 & 14 \\
    \midrule
    Llama3.2-3B & 1.0 & $\attnscore$ & 24 & 25 & 12 & 0 & 24 & 21 & 14 \\
    Llama3.2-3B & 1.0 & $\attnlogdet$ & 12 & 18 & 26 & 23 & 25 & 25 & 12 \\
    Llama3.2-3B & 1.0 & $\attneig$ & 11 & 14 & 27 & 25 & 25 & 27 & 10 \\
    Llama3.2-3B & 1.0 & $\lapeig$ & 11 & 10 & 18 & 12 & 25 & 25 & 11 \\
    \midrule
    Phi3.5 & 0.1 & $\attnscore$ & 7 & 1 & 15 & 0 & 0 & 0 & 19 \\
    Phi3.5 & 0.1 & $\attnlogdet$ & 20 & 19 & 18 & 16 & 17 & 13 & 23 \\
    Phi3.5 & 0.1 & $\attneig$ & 18 & 18 & 19 & 15 & 19 & 18 & 28 \\
    Phi3.5 & 0.1 & $\lapeig$ & 18 & 23 & 28 & 28 & 19 & 31 & 28 \\
    \midrule
    Phi3.5 & 1.0 & $\attnscore$ & 19 & 1 & 0 & 1 & 0 & 0 & 19 \\
    Phi3.5 & 1.0 & $\attnlogdet$ & 12 & 19 & 29 & 14 & 19 & 13 & 14 \\
    Phi3.5 & 1.0 & $\attneig$ & 18 & 1 & 30 & 17 & 31 & 31 & 31 \\
    Phi3.5 & 1.0 & $\lapeig$ & 18 & 16 & 28 & 15 & 19 & 31 & 31 \\
    \midrule
    Mistral-Nemo & 0.1 & $\attnscore$ & 2 & 27 & 18 & 35 & 0 & 30 & 35 \\
    Mistral-Nemo & 0.1 & $\attnlogdet$ & 37 & 20 & 17 & 15 & 38 & 38 & 33 \\
    Mistral-Nemo & 0.1 & $\attneig$ & 38 & 37 & 38 & 18 & 18 & 15 & 31 \\
    Mistral-Nemo & 0.1 & $\lapeig$ & 16 & 38 & 37 & 37 & 18 & 37 & 8 \\
    \midrule
    Mistral-Nemo & 1.0 & $\attnscore$ & 10 & 2 & 16 & 28 & 14 & 30 & 21 \\
    Mistral-Nemo & 1.0 & $\attnlogdet$ & 18 & 17 & 20 & 18 & 18 & 15 & 18 \\
    Mistral-Nemo & 1.0 & $\attneig$ & 38 & 30 & 39 & 39 & 18 & 15 & 18 \\
    Mistral-Nemo & 1.0 & $\lapeig$ & 16 & 39 & 37 & 37 & 18 & 37 & 18 \\
    \midrule
    Mistral-Small-24B & 0.1 & $\attnscore$ & 14 & 1 & 39 & 33 & 35 & 0 & 30 \\
    Mistral-Small-24B & 0.1 & $\attnlogdet$ & 16 & 29 & 38 & 18 & 16 & 38 & 11 \\
    Mistral-Small-24B & 0.1 & $\attneig$ & 36 & 27 & 36 & 19 & 16 & 38 & 20 \\
    Mistral-Small-24B & 0.1 & $\lapeig$ & 21 & 3 & 35 & 24 & 36 & 35 & 34 \\
    \midrule
    Mistral-Small-24B & 1.0 & $\attnscore$ & 15 & 1 & 1 & 0 & 1 & 0 & 30 \\
    Mistral-Small-24B & 1.0 & $\attnlogdet$ & 14 & 24 & 27 & 17 & 24 & 38 & 34 \\
    Mistral-Small-24B & 1.0 & $\attneig$ & 36 & 39 & 27 & 21 & 24 & 36 & 23 \\
    Mistral-Small-24B & 1.0 & $\lapeig$ & 21 & 39 & 36 & 16 & 21 & 35 & 34 \\
    \bottomrule
\end{tabular}

    }
    \label{tab:best_layer_idx}
\end{table*}

\subsection{Comparison with hidden-states-based baselines}
\label{sec:appendix_hidden_states}
We take an approach considered in the previous works \cite{azaria_internal_2023, orgad_llms_2025} and aligned to our evaluation protocol. Specifically, we trained a logistic regression classifier on PCA-projected hidden states to predict whether the model is hallucinating or not. To this end, we select the last token of the answer. While we also tested the last token of the prompt, we observed significantly lower performance, which aligns with results presented by \citep{orgad_llms_2025}. We considered hidden states either from all layers or a single layer corresponding to the selected token. In the \textit{all-layer} scenario, we use the concatenation of hidden states of all layers, and in the \textit{per-layer} scenario, we use the hidden states of each layer separately and select the best-performing layer.

In Table \ref{tab:hidden_states}, we show the obtained results. The \textit{all-layer} version is consistently worse than our $\lapeig$, which further confirms the strength of the proposed method. Our work is one of the first to detect hallucinations solely using attention maps, providing an important insight into the behavior of LLMs, and it motivates further theoretical research on information flow patterns inside these models.

\begin{table*}[htb]
    \centering
    \caption{Results of the probe trained on the hidden state features from the last generated token.}
    \small
    \resizebox{\textwidth}{!}{
    \begin{tabular}{llllllrrrrrr}
     \toprule
     LLM & Temp & Features & \textit{per-layer} & \textit{all-layers} & \multicolumn{7}{c}{Test AUROC ($\uparrow$)} \\
     \cmidrule(lr){6-12}
     \multicolumn{5}{c}{} & CoQA & GSM8K & HaluevalQA & NQOpen & SQuADv2 & TriviaQA & TruthfulQA \\
    \midrule
    Llama3.1-8B & 0.1 & $\hiddenstate$ & \checkmark &  & 0.835 & 0.799 & 0.840 & 0.766 & 0.736 & 0.820 & \textbf{0.834} \\
    Llama3.1-8B & 0.1 & $\hiddenstate$ &  & \checkmark & 0.821 & 0.765 & 0.825 & 0.728 & 0.723 & 0.791 & 0.785 \\
    Llama3.1-8B & 0.1 & $\lapeig$ & \checkmark &  & 0.757 & 0.844 & 0.793 & 0.711 & 0.733 & 0.780 & 0.764 \\
    Llama3.1-8B & 0.1 & $\lapeig$ &  & \checkmark & \textbf{0.836} & \textbf{0.887} & \textbf{0.867} & \textbf{0.793} & \textbf{0.782} & \textbf{0.872} & 0.822 \\
    \midrule
    Llama3.1-8B & 1.0 & $\hiddenstate$ & \checkmark &  & \textbf{0.836} & 0.816 & 0.850 & 0.786 & 0.754 & 0.850 & 0.823 \\
    Llama3.1-8B & 1.0 & $\hiddenstate$ &  & \checkmark & 0.835 & 0.759 & 0.847 & 0.757 & 0.749 & 0.838 & 0.808 \\
    Llama3.1-8B & 1.0 & $\lapeig$ & \checkmark &  & 0.743 & 0.833 & 0.789 & 0.725 & 0.724 & 0.794 & 0.764 \\
    Llama3.1-8B & 1.0 & $\lapeig$ &  & \checkmark & 0.830 & \textbf{0.872} & \textbf{0.874} & \textbf{0.827} & \textbf{0.791} & \textbf{0.889} & \textbf{0.829} \\
    \midrule
    Llama3.2-3B & 0.1 & $\hiddenstate$ & \checkmark &  & 0.800 & 0.826 & 0.808 & 0.732 & 0.750 & 0.782 & 0.760 \\
    Llama3.2-3B & 0.1 & $\hiddenstate$ &  & \checkmark & 0.790 & 0.802 & 0.784 & 0.709 & 0.721 & 0.760 & \textbf{0.770} \\
    Llama3.2-3B & 0.1 & $\lapeig$ & \checkmark &  & 0.676 & 0.835 & 0.774 & 0.730 & 0.727 & 0.712 & 0.690 \\
    Llama3.2-3B & 0.1 & $\lapeig$ &  & \checkmark & \textbf{0.801} & \textbf{0.852} & \textbf{0.844} & \textbf{0.771} & \textbf{0.778} & \textbf{0.821} & 0.743 \\
    \midrule
    Llama3.2-3B & 1.0 & $\hiddenstate$ & \checkmark &  & 0.778 & 0.727 & 0.758 & 0.679 & 0.719 & 0.773 & 0.716 \\
    Llama3.2-3B & 1.0 & $\hiddenstate$ &  & \checkmark & 0.773 & 0.652  & 0.753 & 0.657 & 0.681 & 0.761 & 0.618 \\
    Llama3.2-3B & 1.0 & $\lapeig$ & \checkmark &  & 0.715 & 0.815 & 0.765 & 0.696 & 0.696 & 0.738 & 0.767 \\
    Llama3.2-3B & 1.0 & $\lapeig$ &  & \checkmark & \textbf{0.812} & \textbf{0.870} & \textbf{0.857} & \textbf{0.798} & \textbf{0.751} & \textbf{0.836} & \textbf{0.787} \\
    \midrule
    Phi3.5 & 0.1 & $\hiddenstate$ & \checkmark &  & \textbf{0.841} & 0.773 & \textbf{0.845} & 0.813 & 0.781 & \textbf{0.886} & 0.737 \\
    Phi3.5 & 0.1 & $\hiddenstate$ &  & \checkmark & 0.833 & 0.696 & 0.840 & 0.806 & 0.774 & 0.878 & 0.689 \\
    Phi3.5 & 0.1 & $\lapeig$ & \checkmark &  & 0.716 & 0.753 & 0.757 & 0.761 & 0.732 & 0.768 & \textbf{0.741} \\
    Phi3.5 & 0.1 & $\lapeig$ &  & \checkmark & 0.810 & \textbf{0.785} & 0.819 & \textbf{0.815} &\textbf{ 0.791} & 0.858 & 0.717 \\
    \midrule
    Phi3.5 & 1.0 & $\hiddenstate$ & \checkmark &  & \textbf{0.872} & 0.784 & \textbf{0.850} & 0.821 & 0.806 & \textbf{0.891} & \textbf{0.822} \\
    Phi3.5 & 1.0 & $\hiddenstate$ &  & \checkmark & 0.853 & 0.686 & 0.844 & 0.804 & 0.790 & 0.887 & 0.752 \\
    Phi3.5 & 1.0 & $\lapeig$ & \checkmark &  & 0.723 & 0.816 & 0.769 & 0.755 & 0.732 & 0.792 & 0.732 \\
    Phi3.5 & 1.0 & $\lapeig$ &  & \checkmark & 0.821 & \textbf{0.885} & 0.836 & \textbf{0.826} & 0.795 & 0.872 & 0.777 \\
    \midrule
    Mistral-Nemo & 0.1 & $\hiddenstate$ & \checkmark &  & 0.818 & 0.757 & 0.814 & 0.734 & 0.731 & 0.821 & \textbf{0.792} \\
    Mistral-Nemo & 0.1 & $\hiddenstate$ &  & \checkmark & 0.805 & 0.741 & 0.784 & 0.722 & 0.730 & 0.793 & 0.699 \\
    Mistral-Nemo & 0.1 & $\lapeig$ & \checkmark &  & 0.759 & 0.751 & 0.760 & 0.697 & 0.696 & 0.769 & 0.710 \\
    Mistral-Nemo & 0.1 & $\lapeig$ &  & \checkmark & \textbf{0.823} & \textbf{0.805} & \textbf{0.821} & \textbf{0.755} & \textbf{0.767} & \textbf{0.858} & 0.737 \\
    \midrule
    Mistral-Nemo & 1.0 & $\hiddenstate$ & \checkmark &  & 0.793 & 0.832 & 0.777 & 0.738 & 0.719 & 0.783 & 0.722 \\
    Mistral-Nemo & 1.0 & $\hiddenstate$ &  & \checkmark & 0.771 & 0.834 & 0.771 & 0.706 & 0.685 & 0.779 & 0.644 \\
    Mistral-Nemo & 1.0 & $\lapeig$ & \checkmark &  & 0.738 & 0.808 & 0.763 & 0.708 & 0.723 & 0.785 & 0.818 \\
    Mistral-Nemo & 1.0 & $\lapeig$ &  & \checkmark & \textbf{0.835} & \textbf{0.890} & \textbf{0.833} & \textbf{0.795} & \textbf{0.812} & \textbf{0.865} & \textbf{0.828} \\
    \midrule
    Mistral-Small-24B & 0.1 & $\hiddenstate$ & \checkmark &  & 0.838 & 0.872 & 0.744 & 0.680 & 0.700 & 0.749 & 0.735 \\
    Mistral-Small-24B & 0.1 & $\hiddenstate$ &  & \checkmark & 0.815 & 0.812 & 0.703 & 0.632 & 0.629 & 0.726 & 0.589 \\
    Mistral-Small-24B & 0.1 & $\lapeig$ & \checkmark &  & 0.800 & 0.850 & 0.719 & 0.674 & 0.784 & 0.757 & \textbf{0.827} \\
    Mistral-Small-24B & 0.1 & $\lapeig$ &  & \checkmark & \textbf{0.852} & \textbf{0.881} & \textbf{0.808} & \textbf{0.722} & \textbf{0.821} & \textbf{0.831} & 0.757 \\
    \midrule
    Mistral-Small-24B & 1.0 & $\hiddenstate$ & \checkmark &  & 0.801 & 0.879 & 0.720 & 0.665 & 0.603 & 0.684 & 0.581 \\
    Mistral-Small-24B & 1.0 & $\hiddenstate$ &  & \checkmark & 0.770 & 0.760 & 0.703 & 0.617 & 0.575 & 0.659 & 0.485 \\
    Mistral-Small-24B & 1.0 & $\lapeig$ & \checkmark &  & 0.805 & 0.897 & 0.790 & 0.712 & 0.781 & 0.779 & 0.725 \\
    Mistral-Small-24B & 1.0 & $\lapeig$ &  & \checkmark & \textbf{0.861} & \textbf{0.925} & \textbf{0.882} & \textbf{0.791} & \textbf{0.820} & \textbf{0.876} & \textbf{0.748} \\
    \bottomrule
    \end{tabular}
    }
    \label{tab:hidden_states}
\end{table*}

\section{Extended results of ablations}
\label{sec:appendix_detailed_ablation}
\begin{figure}[!htb]
    \centering
    \includegraphics[width=\columnwidth]{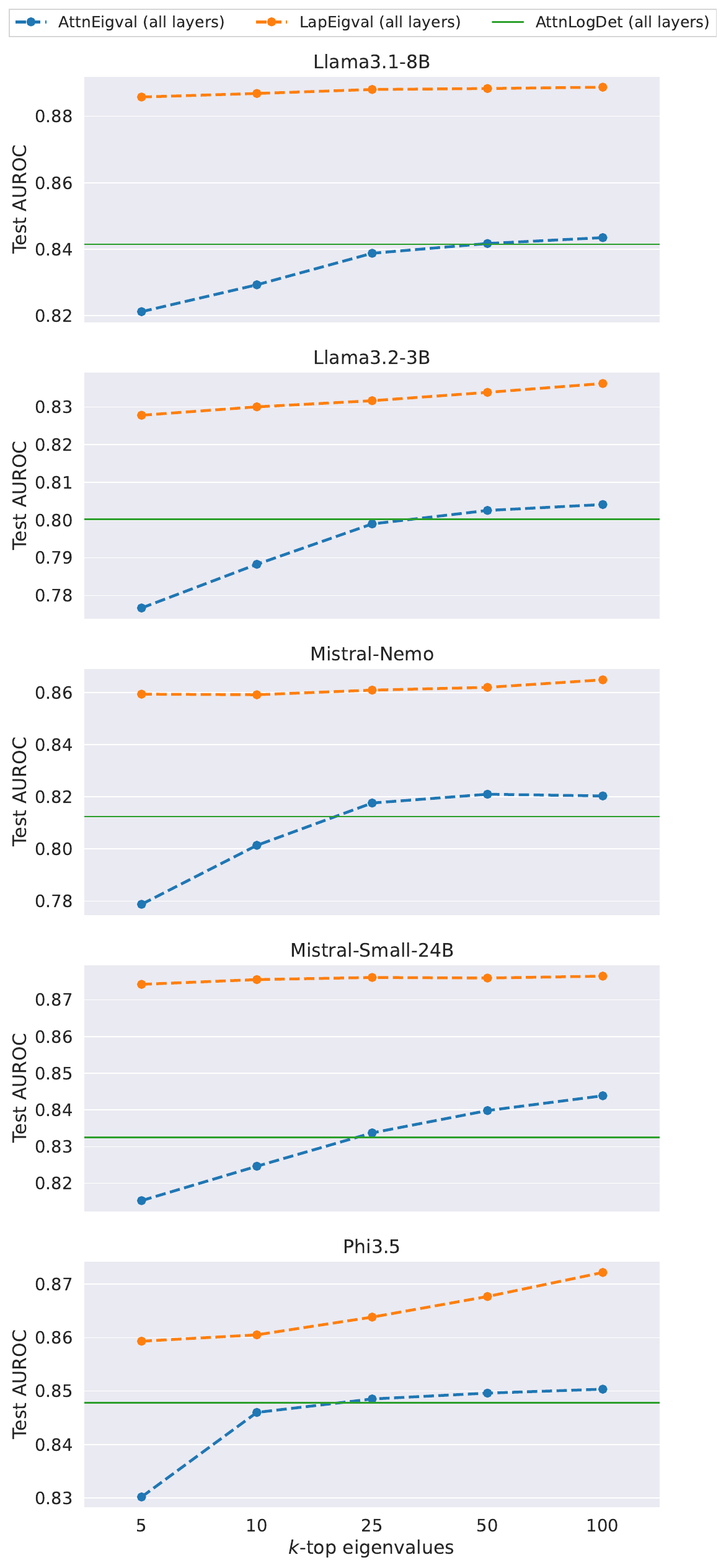}
    \caption{Probe performance across different top-$k$ eigenvalues: $k \in \{5, 10, 25, 50, 100\}$ for TriviaQA dataset with $temp{=}1.0$ and five considered LLMs.}
    \label{fig:appendix_top_k_ablation}
\end{figure}

\begin{figure}[!htb]
    \centering
    \includegraphics[width=\columnwidth]{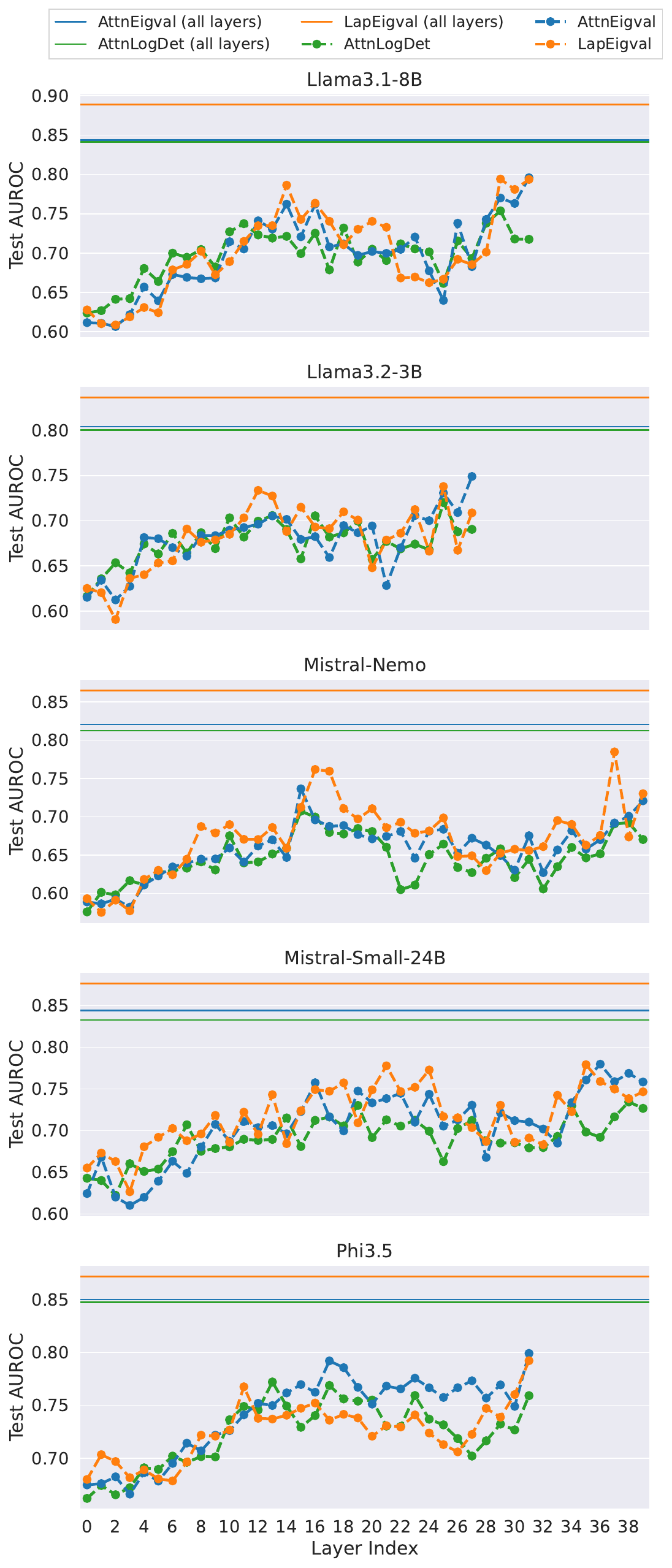}
    \caption{Analysis of model performance across different layers for and 5 considered LLMs and TriviaQA dataset with $temp{=}1.0$ and $k{=}100$ top eigenvalues (results for models operating on all layers provided for reference).}
    \label{fig:appendix_layer_idx_ablation}
\end{figure}

In the following section, we extend the ablation results presented in Section \ref{sec:ablation_topk} and Section \ref{sec:ablation_layer}. Figure \ref{fig:appendix_top_k_ablation} compares the top $k$ eigenvalues across all five LLMs. In  Figure \ref{fig:appendix_layer_idx_ablation} we present a layer-wise performance comparison for each model. 

\section{Extended results of generalization study}
\label{sec:appendix_generalization}
\begin{figure*}[htb]
    \centering
    \includegraphics[width=\textwidth]{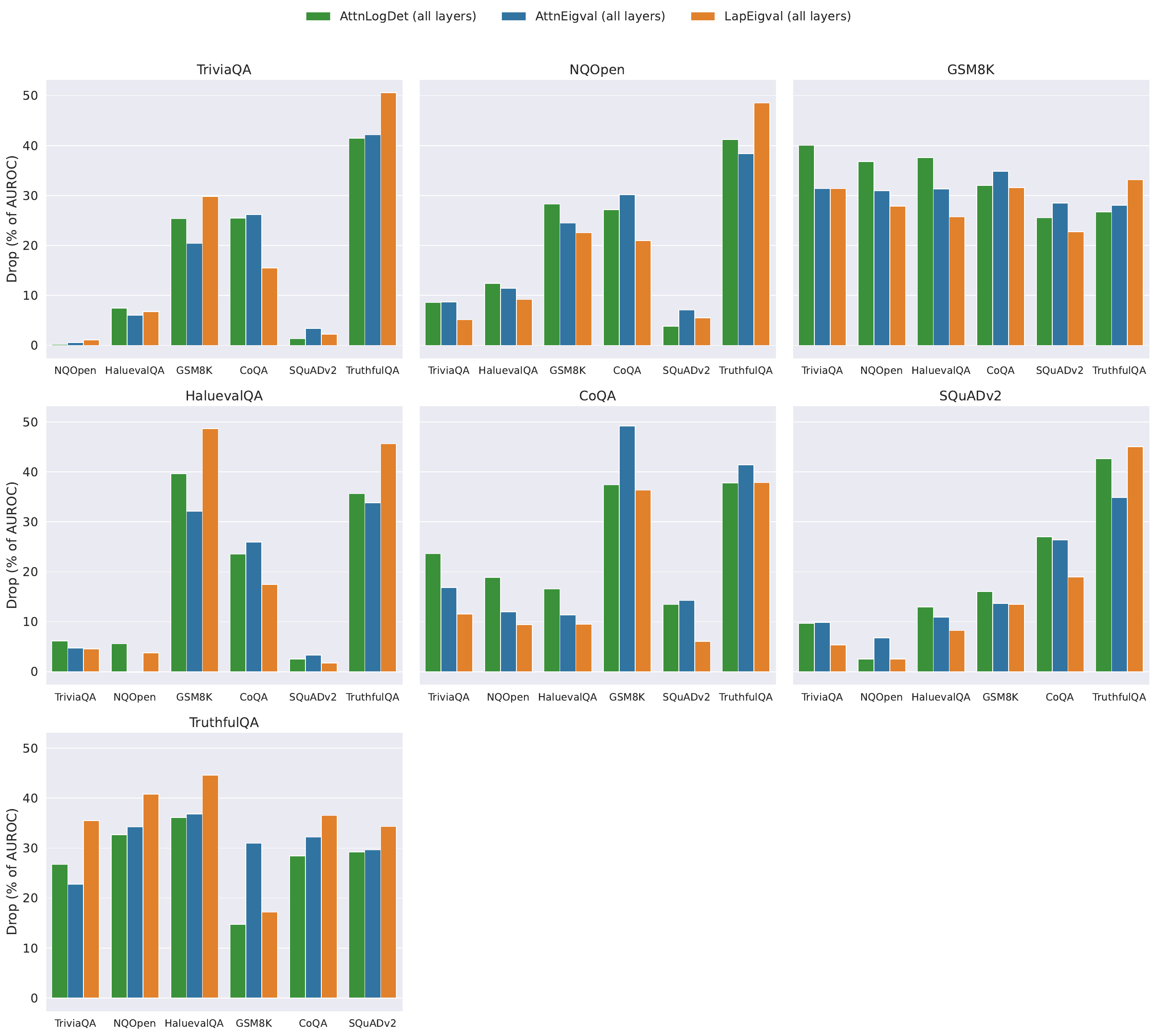}
    \caption{Generalization across datasets measured as a percent performance drop in Test AUROC (less is better) when trained on one dataset and tested on the other. Training datasets are indicated in the plot titles, while test datasets are shown on the $x$-axis. Results computed on \llamabig\ with $k{=}100$ top eigenvalues and $temp{=}1.0$.}
    \label{fig:extended_generalization}
\end{figure*}

We present the complete results of the generalization ablation discussed in Section~\ref{sec:generalization} of the main paper. Table~\ref{tab:extended_generalization} reports the absolute Test AUROC values for each method and test dataset. Except for TruthfulQA, $\lapeig$ achieves the highest performance across all configurations. Notably, some methods perform close to random, whereas $\lapeig$ consistently outperforms this baseline. Regarding relative performance drop (Figure~\ref{fig:extended_generalization}), $\lapeig$ remains competitive, exhibiting the lowest drop in nearly half of the scenarios. These results indicate that our method is robust but warrants further investigation across more datasets, particularly with a deeper analysis of TruthfulQA.

\begin{table*}[htb]
    \centering
    \caption{Full results of the generalization study. By \textcolor{gray}{gray} color we denote results obtained on test split from the same QA dataset as training split, otherwise results are from test split of different QA dataset. We highlight the best performance in \textbf{bold}.}
    \small
    \begin{tabular}{llrrrrrrr}
    \toprule
    Feature & Train Dataset & \multicolumn{7}{c}{Test AUROC ($\uparrow$)} \\
    \cmidrule(lr){3-9}
    & & CoQA & GSM8K & HaluevalQA & NQOpen & SQuADv2 & TriviaQA & TruthfulQA \\
    \midrule
    $\attnlogdet$ & CoQA & \textcolor{gray}{0.758} & 0.518 & 0.687 & 0.644 & 0.646 & 0.640 & 0.587 \\
    $\attneig$ & CoQA & \textcolor{gray}{0.782} & 0.426 & 0.726 & 0.696 & 0.659 & 0.702 & 0.560 \\
    $\lapeig$ & CoQA & \textcolor{gray}{0.830} & \textbf{0.555} & \textbf{0.790} & \textbf{0.748} & \textbf{0.743} & \textbf{0.786} & \textbf{0.629} \\
    \midrule
    $\attnlogdet$ & GSM8K & 0.515 & \textcolor{gray}{0.828} & 0.513 & 0.502 & 0.555 & 0.503 & \textbf{0.586} \\
    $\attneig$ & GSM8K & 0.510 & \textcolor{gray}{0.838} & 0.563 & 0.545 & 0.549 & 0.579 & 0.557 \\
    $\lapeig$ & GSM8K & \textbf{0.568} & \textcolor{gray}{0.872} & \textbf{0.648} & \textbf{0.596} & \textbf{0.611} & \textbf{0.610} & 0.538 \\
    \midrule
    $\attnlogdet$ & HaluevalQA & 0.580 & 0.500 & \textcolor{gray}{0.823} & 0.750 & 0.727 & 0.787 & 0.668 \\
    $\attneig$ & HaluevalQA & 0.579 & \textbf{0.569} & \textcolor{gray}{0.819} & 0.792 & 0.743 & 0.803 & \textbf{0.688} \\
    $\lapeig$ & HaluevalQA & \textbf{0.685} & 0.448 & \textcolor{gray}{0.873} & \textbf{0.796} & \textbf{0.778} & \textbf{0.848} & 0.595 \\
    \midrule
    $\attnlogdet$ & NQOpen & 0.552 & 0.594 & 0.720 & \textcolor{gray}{0.794} & 0.717 & 0.766 & 0.597 \\
    $\attneig$ & NQOpen & 0.546 & 0.633 & 0.725 & \textcolor{gray}{0.790} & 0.714 & 0.770 & \textbf{0.618} \\
    $\lapeig$ & NQOpen & \textbf{0.656} & \textbf{0.676} & \textbf{0.792} & \textcolor{gray}{0.827} & \textbf{0.748} & \textbf{0.843} & 0.564 \\
    \midrule
    $\attnlogdet$ & SQuADv2 & 0.553 & 0.695 & 0.716 & 0.774 & \textcolor{gray}{0.746} & 0.757 & 0.658 \\
    $\attneig$ & SQuADv2 & 0.576 & 0.723 & 0.730 & 0.737 & \textcolor{gray}{0.768} & 0.760 & \textbf{0.711} \\
    $\lapeig$ & SQuADv2 & \textbf{0.673} & \textbf{0.754} & \textbf{0.801} & \textbf{0.806} & \textcolor{gray}{0.791} & \textbf{0.841} & 0.625 \\
    \midrule
    $\attnlogdet$ & TriviaQA & 0.565 & 0.618 & 0.761 & 0.793 & 0.736 & \textcolor{gray}{0.838} & 0.572 \\
    $\attneig$ & TriviaQA & 0.577 & \textbf{0.667} & 0.770 & 0.786 & 0.742 & \textcolor{gray}{0.843} & \textbf{0.616} \\
    $\lapeig$ & TriviaQA & \textbf{0.702} & 0.612 & \textbf{0.813} & \textbf{0.818} & \textbf{0.773} & \textcolor{gray}{0.889} & 0.522 \\
    \midrule
    $\attnlogdet$ & TruthfulQA & 0.550 & 0.706 & 0.597 & \textbf{0.603} & 0.604 & 0.662 & \textcolor{gray}{0.811} \\
    $\attneig$ & TruthfulQA & 0.538 & 0.579 & \textbf{0.600} & 0.595 & \textbf{0.646} & \textbf{0.685} & \textcolor{gray}{0.833} \\
    $\lapeig$ & TruthfulQA & \textbf{0.590} & \textbf{0.722} & 0.552 & 0.529 & 0.569 & 0.631 & \textcolor{gray}{0.829} \\
    \bottomrule
    \end{tabular}
    \label{tab:extended_generalization}
\end{table*}

\section{Influence of dataset size}
One of the limitations of $\lapeig$ is that it is a supervised method and thus requires labelled hallucination data. To check whether it requires a large volume of data, we conducted an additional study in which we trained $\lapeig$ on only a stratified fraction of the available examples for each hallucination dataset (using a dataset created from $\llamabig$ outputs) and evaluated on the full test split. The AUROC scores are presented in Table \ref{tab:dataset_fraction_influence}. As shown, LapEigvals maintains reasonable performance even when trained on as few as a few hundred examples. Additionally, we emphasise that labelling can be efficiently automated and scaled using the $\llmjudge$ paradigm.

\begin{table*}[ht]
    \centering
    \caption{Impact of training dataset size on performance. Test AUROC scores are reported for different fractions of the training data. The study uses a dataset derived from $\llamabig$ answers with $temp{=}1.0$ and $k{=}100$ top eigenvalues, with absolute dataset sizes shown in parentheses.}
    \resizebox{\textwidth}{!}{%
        \begin{tabular}{lccccccc}
\toprule
Fraction of data (\%) & CoQA (6316) & GSM8K (1019) & HaluEvalQA (6118) & NQOpen (2360) & SQuADv2 (3818) & TriviaQA (7710) & TruthfulQA (596) \\
\midrule
100 & 0.830 & 0.872 & 0.873 & 0.827 & 0.791 & 0.889 & 0.804 \\
75 & 0.824 & 0.867 & 0.868 & 0.816 & 0.785 & 0.886 & 0.803 \\
50 & 0.817 & 0.858 & 0.861 & 0.802 & 0.778 & 0.880 & 0.796 \\
30 & 0.802 & 0.851 & 0.853 & 0.785 & 0.760 & 0.872 & 0.786 \\
20 & 0.790 & 0.835 & 0.848 & 0.770 & 0.738 & 0.863 & 0.763 \\
10 & 0.757 & 0.816 & 0.829 & 0.726 & 0.730 & 0.841 & 0.709 \\
5 & 0.734 & 0.764 & 0.811 & 0.668 & 0.702 & 0.813 & 0.637 \\
1 & 0.612 & 0.695 & 0.736 & 0.621 & 0.605 & 0.670 & 0.545 \\

\bottomrule
\end{tabular}
    }
    \label{tab:dataset_fraction_influence}
\end{table*}

\section{Reliability of spectral features}
Our method relies on ordered spectral features, which may exhibit sensitivity to perturbations and limited robustness. In our setup, both attention weights and extracted features were stored as \texttt{bfloat16} type, which has lower precision than \texttt{float32}. The reduced precision acts as a form of regularization--minor fluctuations are often rounded off, making the method more robust to small perturbations that might otherwise affect the eigenvalue ordering.

To further investigate perturbation-sensitivity, we conducted a controlled analysis on one model by adding Gaussian noise to randomly selected input feature dimensions before the eigenvalue sorting step. We varied both the noise standard deviation and the fraction of perturbed dimensions (ranging from 0.5 to 1.0). Perturbations were applied consistently to both the training and test sets. In Table \ref{tab:perturbation_results} we report the mean and standard deviation of performance across 5 runs on hallucination data generated by $\llamabig$ on the TriviaQA dataset with $temp{=}1.0$, along with percentage change relative to the unperturbed baseline (0.0 indicates no perturbation applied). We observe that small perturbations have a negligible impact on performance and further confirm the robustness of our method.

\begin{table*}[ht]
    \centering
    \caption{Impact of Gaussian noise perturbations on input features for different top-$k$ eigenvalues and noise standard deviations $\sigma$. Results are averaged over five perturbations, with mean and standard deviation reported; relative percentage drops are shown in parentheses. Results were obtained for $\llamabig$ with $temp{=}1.0$ on TriviaQA dataset.}
    \resizebox{\textwidth}{!}{%
        \begin{tabular}{lcccccc}
\toprule
\textbf{$k$} & \multicolumn{6}{c}{Test AUROC ($\uparrow$)} \\
\cmidrule(lr){2-7}
           & $\sigma=0.0$ & $\sigma=1e-5$ & $\sigma=1e-4$ & $\sigma=1e-3$ & $\sigma=1e-2$ & $\sigma=1e-1$ \\
\midrule
5 & 0.867 ± 0.0 (0.0\%) & 0.867 ± 0.0 (0.0\%) & 0.867 ± 0.0 (0.0\%) & 0.867 ± 0.0 (-0.01\%) & 0.859 ± 0.003 (0.86\%) & 0.573 ± 0.017 (33.84\%) \\
10 & 0.867 ± 0.0 (0.0\%) & 0.867 ± 0.0 (0.0\%) & 0.867 ± 0.0 (0.0\%) & 0.867 ± 0.0 (0.03\%) & 0.861 ± 0.002 (0.78\%) & 0.579 ± 0.01 (33.3\%) \\
20 & 0.869 ± 0.0 (0.0\%) & 0.869 ± 0.0 (0.0\%) & 0.869 ± 0.0 (0.0\%) & 0.869 ± 0.0 (0.0\%) & 0.862 ± 0.002 (0.84\%) & 0.584 ± 0.018 (32.76\%) \\
50 & 0.870 ± 0.0 (0.0\%) & 0.870 ± 0.0 (0.0\%) & 0.870 ± 0.0 (0.0\%) & 0.869 ± 0.0 (0.02\%) & 0.864 ± 0.002 (0.66\%) & 0.606 ± 0.014 (30.31\%) \\
100 & 0.872 ± 0.0 (0.0\%) & 0.872 ± 0.0 (0.0\%) & 0.872 ± 0.0 (0.01\%) & 0.872 ± 0.0 (-0.0\%) & 0.866 ± 0.001 (0.66\%) & 0.640 ± 0.007 (26.64\%) \\
\bottomrule
\end{tabular}
    }
    \label{tab:perturbation_results}
\end{table*}

\section{Cost and time analysis}
Providing precise cost and time measurements is nontrivial due to the multi-stage nature of our method, as it involves external services (e.g., OpenAI API for labelling), and the runtime and cost can vary depending on the hardware and platform used. Nonetheless, we present an overview of the costs and complexity as follows.

\begin{enumerate}
    \item Inference with LLM (preparing hallucination dataset) - does not introduce additional cost beyond regular LLM inference; however, it may limit certain optimizations (e.g. FlashAttention \citep{10.5555/3600270.3601459}) since the full attention matrix needs to be materialized in memory.
    
    \item Automated labeling with $\llmjudge$ using OpenAI API - we estimate labeling costs using the \texttt{tiktoken} library and OpenAI API pricing (\$0.60 per 1M output tokens). However, these estimates exclude caching effects and could be reduced using the Batch API; Table \ref{tab:cost_estimation} reports total and per-item hallucination labelling costs across all datasets (including 5 LLMs and 2 temperature settings). Estimation for GSM8K dataset is not present as the outputs for this dataset are evaluated by exact-match.
    
    \item Computing spectral features - since we exploit the fact that eigenvalues of the Laplacian lie on the diagonal, the complexity is dominated by the computation of the out-degree matrix, which in turn is dominated by the computation of the mean over rows of the attention matrix. Thus, it is $O(n^2)$ time, where $n$ is the number of tokens. Then, we have to sort eigenvalues, which takes $O(n \log n)$ time. The overall complexity multiplies by the number of layers and heads of a particular LLM. Practically, in our implementation, we fused feature computation with LLM inference, since we observed a memory bottleneck compared to using raw attention matrices stored on disk.
\end{enumerate}

\begin{table*}[ht]
    \centering
    \caption{Estimation of costs regarding $\llmjudge$ labelling with OpenAI API.}
    \resizebox{\textwidth}{!}{
    \begin{tabular}{lrrrrrrr}
\toprule
Dataset & Total Input Tokens & Total Output Tokens & Mean Input Tokens & Mean Output Tokens & Total Input Cost [\$] & Total Output Cost [\$] & Total Cost [\$] \\
\midrule
CoQA       & 52,194,357  & 320,613  & 653.82 & 4.02 & 7.83 & 0.19 & 8.02 \\
NQOpen     & 11,853,621  & 150,782  & 328.36 & 4.18 & 1.78 & 0.09 & 1.87 \\
HaluEvalQA & 33,511,346  & 421,572  & 335.11 & 4.22 & 5.03 & 0.25 & 5.28 \\
SQuADv2    & 19,601,322  & 251,264  & 330.66 & 4.24 & 2.94 & 0.15 & 3.09 \\
TriviaQA   & 41,114,137  & 408,067  & 412.79 & 4.10 & 6.17 & 0.24 & 6.41 \\
TruthfulQA &  2,908,183  &  33,836  & 355.96 & 4.14 & 0.44 & 0.02 & 0.46 \\
\midrule
\textbf{Total} & \textbf{158,242,166} & \textbf{1,575,134} & \textbf{402.62} & \textbf{4.15} & \textbf{24.19} & \textbf{0.94} & \textbf{25.13} \\
\bottomrule
\end{tabular}
    }
    \label{tab:cost_estimation}
\end{table*}

\section{QA prompts}
\label{sec:appendix_prompts}
Following, we describe all prompts for QA used to obtain the results presented in this work:
\begin{itemize}
    \item prompt $p_1$ -- medium-length one-shot prompt with single example of QA task (Listing~\ref{lst:p1}),
    \item prompt $p_2$ -- medium-length zero-shot prompt without examples (Listing~\ref{lst:p2}),
    \item prompt $p_3$ -- long few-shot prompt; the main prompt used in this work; modification of prompt used by \citep{kossen_semantic_2024} (Listing~\ref{lst:p3}),
    \item prompt $p_4$ -- short-length zero-shot prompt without examples (Listing~\ref{lst:p4}),
    \item prompt $gsm8k$ -- short prompt used for GSM8K dataset with output-format instruction.
\end{itemize}

\begin{figure*}[htb]
\centering
\begin{lstlisting}[style=prompt, caption={One-shot QA (prompt $p_1$)}, label=lst:p1]
Deliver a succinct and straightforward answer to the question below. Focus on being brief while maintaining essential information. Keep extra details to a minimum.

Here is an example:
Question: What is the Riemann hypothesis?
Answer: All non-trivial zeros of the Riemann zeta function have real part 1/2

Question: {question}
Answer:
\end{lstlisting}
\end{figure*}

\begin{figure*}[htb]
\centering
\begin{lstlisting}[style=prompt, caption={Zero-shot QA (prompt $p_2$).}, label=lst:p2]
Please provide a concise and direct response to the following question, keeping your answer as brief and to-the-point as possible while ensuring clarity. Avoid any unnecessary elaboration or additional details.
Question: {question}
Answer:
\end{lstlisting}
\end{figure*}

\begin{figure*}[htb]
\centering
\begin{lstlisting}[style=prompt, mathescape, caption={Few-shot QA prompt (prompt $p_3$), modified version of prompt used by \citep{kossen_semantic_2024}.}, label=lst:p3]
Answer the following question as briefly as possible.
Here are several examples:
Question: What is the capital of France?
Answer: Paris

Question: Who wrote *Romeo and Juliet*?
Answer: William Shakespeare

Question: What is the boiling point of water in Celsius?
Answer: 100${^\circ}$C

Question: How many continents are there on Earth?
Answer: Seven

Question: What is the fastest land animal?
Answer: Cheetah

Question: {question}
Answer:
\end{lstlisting}
\end{figure*}

\begin{figure*}[htb]
\centering
\begin{lstlisting}[style=prompt, caption={Zero-shot short QA prompt (prompt $p_4$).}, label=lst:p4]
Answer the following question as briefly as possible.
Question: {question}
Answer:
\end{lstlisting}
\end{figure*}

\begin{figure*}[htb]
\centering
\begin{lstlisting}[style=prompt, caption={Zero-shot QA prompt for GSM8K dataset.}, label=lst:gsm8k]
Given the following problem, reason and give a final answer to the problem.

Problem: {question}
Your response should end with "The final answer is [answer]" where [answer] is the response to the problem.
\end{lstlisting}
\end{figure*}

\section{LLM-as-Judge prompt}
During hallucinations dataset construction we leveraged \llmjudge\ approach to label answers generated by the LLMs. To this end, we utilized \texttt{gpt-4o-mini} with prompt in Listing~\ref{lst:judge_prompt}, which is an adapted version of the prompt used by \citep{orgad_llms_2025}.

\begin{figure*}[htb]
\centering
\begin{lstlisting}[style=prompt, caption={Prompt used in \llmjudge\ approach for determining hallucination labels. Prompt is a modified version of the one used by \citep{orgad_llms_2025}.}, label=lst:judge_prompt]
You will evaluate answers to questions. For each question, I will provide a model's answer and one or more correct reference answers.
You would have to determine if the model answer is correct, incorrect, or model refused to answer. The model answer to be correct has to match from one to all of the possible correct answers.
If the model answer is correct, write 'correct' and if it is not correct, write 'incorrect'. If the Model Answer is a refusal, stating that they don't have enough information, write 'refuse'.
For example:

Question: who is the young guitarist who played with buddy guy?
Ground Truth: [Quinn Sullivan, Eric Gales]
Model Answer: Ronnie Earl
Correctness: incorrect

Question: What is the name of the actor who plays Iron Man in the Marvel movies?
Ground Truth: [Robert Downey Jr.]
Model Answer: Robert Downey Jr. played the role of Tony Stark/Iron Man in the Marvel Cinematic Universe films.
Correctness: correct

Question: what is the capital of France?
Ground Truth: [Paris]
Model Answer: I don't have enough information to answer this question.
Correctness: refuse

Question: who was the first person to walk on the moon?
Ground Truth: [Neil Armstrong]
Model Answer: I apologize, but I cannot provide an answer without verifying the historical facts.
Correctness: refuse

Question: {{question}}
Ground Truth: {{gold_answer}}
Model Answer: {{predicted_answer}}
Correctness:
\end{lstlisting}
\end{figure*}

\end{document}